\documentclass[journal]{IEEEtran}

\usepackage{url}
\usepackage{times}
\usepackage{epsfig}
\usepackage{graphicx}
\usepackage{amsmath}
\usepackage{amssymb}
\usepackage{bm}
\usepackage{xspace}
\usepackage[linesnumbered,ruled]{algorithm2e}
\usepackage{booktabs}
\usepackage{multirow}
\usepackage{tabulary}
\usepackage[table]{xcolor}
\usepackage{hyperref}
\hypersetup{
	colorlinks=true,   
	urlcolor=blue
}

\newcolumntype{L}[1]{>{\raggedright\arraybackslash}p{#1}}
\newcolumntype{C}[1]{>{\centering\arraybackslash}p{#1}}
\newcolumntype{R}[1]{>{\raggedleft\arraybackslash}p{#1}}

\newcommand{\etal}{\textit{et al}. }
\newcommand{\ie}{\textit{i}.\textit{e}. }
\newcommand{\eg}{\textit{e}.\textit{g}. }
\newcommand{\fig}{{Figure}\@\xspace}
\newcommand{\tab}{{Table}\@\xspace}
\newcommand{\eqn}{{Eqn.}\@\xspace}
\newcommand{\alg}{{Algorithm}\@\xspace}

\def\Mat#1{{\bm{#1}}}

\graphicspath{{figures/}}

\begin{document}

\title
	{
	Unsupervised Online Video Object Segmentation with Motion Property Understanding
	}

\author
	{Tao~Zhuo,
	 Zhiyong Cheng,
	 Peng~Zhang, 
	 Yongkang~Wong,~\IEEEmembership{Member,~IEEE}
	 and Mohan~Kankanhalli,~\IEEEmembership{Fellow,~IEEE}%
		\thanks{Manuscript received July 20, 2018; revised January 10, 2019, March 30, 2019 and May 17, 2019; accepted July 9, 2019. This research is supported by the National Research Foundation, Prime Minister's Office, Singapore under its Strategic Capability Research Centres Funding Initiative. This research is also supported by National Natural Science Foundation of China 61571362 and Natural Science Basic Research Plan in Shaanxi Province of China (Program No. 2018JM6015). \emph{(Corresponding author: Tao Zhuo)}}
		\thanks{T.~Zhuo, Y.~Wong and M.~Kankanhalli are with the School of Computing, National University of Singapore, Singapore (email: \{zhuotao,yongkang.wong\}@nus.edu.sg and mohan@comp.nus.edu.sg).}
		\thanks{Z.~Cheng is with the Qilu University of Technology (Shandong Academy of Sciences), Shandong Computer Science Center (National Supercomputer Center in Jinan), Shandong Artificial Intelligence Institute, China, (email: jason.zy.cheng@gmail.com).}
		\thanks{P.~Zhang is with the School of Computer Science, Northwestern Polytechnical University, China (email: zh0036ng@nwpu.edu.cn).}
	}

%\markboth{IEEE Transactions on Image Processing,~Vol.~XX, No.~X, June~2018}%
%\markboth{}%

\maketitle

% \fi
\IEEEpeerreviewmaketitle

\begin{abstract}

Unsupervised video object segmentation aims to automatically segment moving objects over an unconstrained video without any user annotation. So far, only few unsupervised online methods have been reported in literature and their performance is still far from satisfactory, because the complementary information from  future frames cannot be processed under online setting. To solve this challenging problem, in this paper, we propose a novel Unsupervised Online Video Object Segmentation (UOVOS) framework by construing the motion property to mean \emph{moving} in concurrence with \emph{a generic object} for segmented regions. By incorporating \emph{salient motion detection} and \emph{object proposal}, a pixel-wise fusion strategy is developed to effectively remove detection noise such as dynamic background and stationary objects. Furthermore, by leveraging the obtained segmentation from immediately preceding frames, a forward propagation algorithm is employed to deal with unreliable motion detection and object proposals. Experimental results on several benchmark datasets demonstrate the efficacy of the proposed method. Compared to the state-of-the-art unsupervised online segmentation algorithms, the proposed method achieves an absolute gain of 6.2\%. Moreover, our method achieves better performance than the best unsupervised offline algorithm on the DAVIS-2016 benchmark dataset. Our code is available on the project website: \url{https://github.com/visiontao/uovos}.

\end{abstract}


\section{Introduction}
\label{sec:introduction}

\IEEEPARstart{T}he task of Video Object Segmentation (VOS) is to separate objects (foreground) from the background.
This is important for the wide range of video understanding applications,
such as video surveillance, unmanned vehicle navigation and action recognition.
Traditionally, most approaches in VOS mainly focused on background modeling in stationary camera scenarios.
Recently, this focus has been shifted from stationary camera to freely moving camera environment \cite{ICCV2013_Papazoglou,TIP2016_Yang,CVPR2014_Liu,TIP2018_Yang,CVPR2016_Perazzi,CVPR2017_Khoreva,CVPR2017_Tokmakov,CVPR2018_Yang,CVPR2018_Li}.
Due to the complex video content (\eg object deformation, background clutter and occlusion) and the dynamic nature of moving background caused by camera motion, moving object segmentation under the moving camera environment is still a challenging problem.

Depending on whether the object mask is manually annotated or not, existing VOS algorithms can be broadly categorized into semi-supervised approach or unsupervised approach. Generally, the former \cite{CVPR2017_Khoreva,SIGGRAPH2015_Fan,CVPR2016_Maerki,CVPR2017_Caelles} aims to segment specific objects based on the user annotation (often the first frame of a video). In contrast, the latter \cite{ICCV2013_Papazoglou,CVPR2017_Tokmakov,CVPR2018_Li,ICCV2009_Sheikh,CVPR2012_Ochs,CVPR2016_Xiao} 
aims to automatically segment moving objects without any user annotation on the given video. In this paper, we mainly focus on the unsupervised VOS task.

The popular unsupervised VOS methods often focus on clustering the long-term trajectories of pixels \cite{CVPR2012_Ochs}, superpixels \cite{ICCV2013_Papazoglou,BMVC2014_Faktor,ECCV2014_Jain} or object proposals \cite{CVPR2016_Xiao,CVPR2015_Fragkiadaki,CVPR2017_Koh} across the entire video,
and the pixels with consistent trajectories are clustered as foreground.
This long-term trajectory-based strategy often requires the entire video sequence upfront to obtain good results.
Thus, it must operate in an offline manner with the following problems.
\begin{enumerate}
	\item The targeted moving object must appear in most frames of the given video \cite{ICCV2013_Papazoglou,CVPR2017_Koh}, otherwise it will probably be classified as background.
	\item The requirement of the entire video implies that the offline methods cannot segment moving objects in a frame-by-frame manner. Therefore, it is impractical for video streaming applications (\eg video surveillance).
	\item Due to the large memory requirement, this strategy also becomes infeasible for analyzing a long video sequence.
\end{enumerate}

\begin{figure*}[!t]
	\centerline{\includegraphics[width=1.0\textwidth]{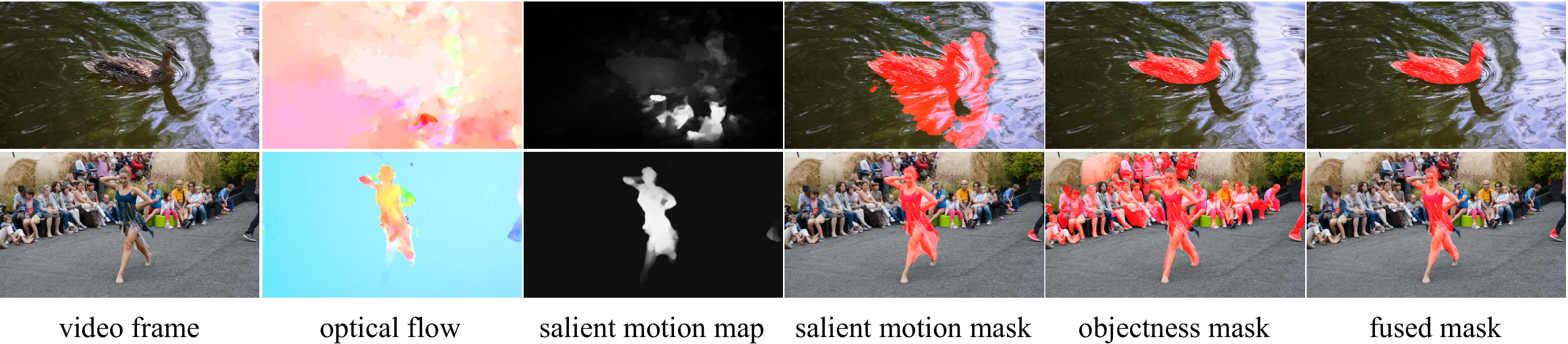}}
	\caption{Two examples of the moving object segmentation with salient motion detection and object proposals. Salient motion map denotes the moving probability of each pixel; salient motion mask represents the extracted moving regions; objectness mask is the detected generic object regions; fused mask is our motion  segmentation result. Based on our fusion method, moving background (\eg moving water) and stationary objects can be effectively removed.}
% 	, as shown in first row and second row, respectively.}
	\label{fig_so}
\end{figure*}

In order to overcome the limitations of offline approaches,
the development of unsupervised online VOS frameworks has attracted more attention.
Wang \etal \cite{CVPR2015_Wang} combined current frame with several forward-backward neighboring frames to generate short-term trajectories.
Based on the spatio-temporal saliency map generated by optical flow field and salient object detection,
moving objects are automatically segmented. However, since a moving object is not always salient in some videos, the spatio-temporal saliency map cannot produce good segmentation results in that case. 
Different from the online strategy with short-term trajectories, 
some researchers adopted another tracking-based unsupervised online framework for VOS.
Briefly,
by automatically initializing the target object on a few frames with different motion cues, 
an online tracking method is then used to propagate the initialized object regions to subsequent frames, 
as in \cite{ICCV2013_Li,CVPR2015_Taylor,ICCV2015_Yang,ECCV2016_Bideau}.
However, the segmentation results are subject to the quality of the initialized object regions.
Besides, these methods suffer from error accumulation~\cite{CVPR2017_Khoreva} when the tracking initialized object regions to the subsequent frames.

Recently, deep learning based methods have been deployed to automatically segment moving objects with motion cues.
For example, Tokmakov \etal \cite{CVPR2017_Tokmakov} adopted an end-to-end framework on the optical flow field for motion segmentation, followed by an object proposals model \cite{ECCV2016_Pinheiro} to extract the candidate objects. Jain \etal \cite{CVPR2017_Jain} proposed a two-stream fully convolutional network to combine the object proposals and motion for segmenting generic objects in videos.
Unlike traditional methods, deep learning based approaches require a large amount of well-annotated data for training.
In addition, when the object movements and video scenarios are very different from the training data,
their performance may degrade substantially.

Based on the above analysis, although much progress has been made by existing methods, developing accurate unsupervised online VOS algorithms remains a challenging problem. In this paper, motivated by the moving object definition in which a segmented region should be \emph{moving} and indicate a \emph{generic object}, which we call \emph{motion property}, we propose a novel fully Unsupervised Online VOS (UOVOS) framework for more accurate moving object segmentation.
To extract the regions that satisfy both moving object properties (\ie \emph{moving} and \emph{generic object}),
we propose a novel motion segmentation method that segments moving objects between two video frames with salient motion detection and object proposals. Specifically, the salient motion detection method is used to extract moving regions (denoted as salient motion mask) on the optical flow; and the object proposals method is applied to detect the generic object regions (denoted as objectness mask) on each frame. 
However, neither the salient motion mask or objectness mask alone can accurately detect regions with both ``moving'' and ``generic objects'' properties. Therefore, we propose a pixel-level fusion method to operate on the intersection of the detected regions by the salient motion map and objectness map. 
As shown in \fig~\ref{fig_so}, by fusing the salient motion detection result and object proposals, the moving background regions and static objects can be effectively removed by our method. Unlike the existing deep learning methods \cite{CVPR2017_Tokmakov,CVPR2017_Jain} that learn the motion segmentation model from a large number of well annotated data, our method does not require any additional training data as it is able to directly employ a pretrained object proposals model \cite{ICCV2017_He} without fine-tuning. 

In addition, due to complex video scenarios, salient motion detection and object proposals in individual frame are not always reliable.
With the observation that the video content in neighboring frames often share consistent motion dynamic, 
% continuously changes,
we propose a forward propagation refinement method to predict more accurate moving and generic object regions. 
By propagating the results of several previous frames to the current frame,
a more accurate segmentation result is estimated with the refined salient motion mask and objectness mask.

Finally, to produce accurate object boundaries,
we adopt a CRF model \cite{NIPS2011_Krahenbuhl} for further segmentation refinement.
Based on the proposed motion segmentation and forward propagation refinement,
our method is able to automatically segment moving objects in an online manner.
To demonstrate the effectiveness of our proposed approach, 
we conduct evaluation on the DAVIS-2016 \cite{CVPR2016_Perazzi}, SegTrack-v2 \cite{ICCV2013_Li} and FBMS-59 \cite{TPAMI2014_Ochs} benchmark datasets.
Experimental results show the effectiveness and competitive accuracy of our method. Besides, compared to the state-of-the-art methods, our method significantly outperforms the unsupervised online algorithms by 6.2\%, and even achieves better performance than the best unsupervised offline on the DAVIS-2016 dataset.

In summary, our main contributions are as follows.
\begin{enumerate}
	\item We propose a novel Unsupervised Online Video Object Segmentation (UOVOS) framework, which utilizes from motion property. In particular, we design a pixel-wise fusion method for the salient motion detection and object proposals, which can effectively remove moving background and stationary object noise.	
	\item To deal with unreliable salient motion and object proposals in complex videos, we propose a forward propagation method by leveraging the segmentation mask from previous frames for mask refinement.
	\item We conduct comprehensive experiments on three benchmark datasets. The experimental results show that our method significantly outperforms the state-of-the-art methods by a large margin of 6.2\%. 
\end{enumerate}

The remainder of the paper is organized as follows.
The related work is reviewed in Section~\ref{sec:relatedwork}.
Section~\ref{sec:proposed} elaborates on our approach. 
Section~\ref{sec:experiment} discusses the experiments and results. 
Section~\ref{sec:conclusion} concludes this work.

\section{Related Work}
\label{sec:relatedwork}

\subsection{Semi-supervised VOS}
Semi-supervised VOS methods aim to segment specific objects in videos based on the user annotation on some video frames (often the first frame of the video). Recent semi-supervised methods \cite{CVPR2017_Caelles,CVPR2016_Tsai,CVPR2015_Wen,BMVC2017_Voigtlaender,CVPR2018_Cheng} often assume that the object mask is known in the first frame, followed by a tracking method to segment it in the subsequent frames.
In order to alleviate the drift problem \cite{TIP2018_Liu} in tracking stage, Fan \etal \cite{SIGGRAPH2015_Fan} annotated the object mask in a few frames, 
and adopted a local mask transfer method to propagate the source annotation to terminal images in both forward and backward directions.
Recently, many deep learning based approaches~\cite{CVPR2017_Khoreva,CVPR2017_Caelles,arxiv2017_Khoreva,CVPR2018_Cheng,BMVC2017_Voigtlaender,CVPR2018_Oh} have been developed for semi-supervised VOS, making significant progress. For example, RGMP method~\cite{CVPR2018_Oh} proposes a hybrid model that fuses the mask detection and propagation in an encoder-decoder network. It can leverage the temporal information from the previous frame and the annotated object mask in the first frame for current frame processing. Benefiting from the effective network architecture design, accurate results can be obtained for both single object and multi-object segmentation by the semi-supervise methods. However, due to the requirement of object annotation in videos, semi-supervised approaches are not feasible in and scalable for some applications, such as video surveillance systems. 

\subsection{Unsupervised VOS}
Unsupervised VOS algorithms aim to automatically segment moving objects without any user annotation. Compared to semi-supervised methods, unsupervised algorithms cannot segment a specific object due to motion ambiguity between different instances and dynamic background.
The early methods \cite{ICCV2009_Sheikh,CVPR2014_Jung} are often based on geometric scene modeling \cite{Book2003_Hartley}, where the geometric model fitting error is used to classify the foreground/background label of the corresponding pixels.
Sheikh \etal \cite{ICCV2009_Sheikh} adopted a homography model to distinguish foreground/background trajectories, 
but they assume an affine model over a more accurate perspective camera model. 
For more accurate scene modeling, Jung \etal \cite{CVPR2014_Jung} used multiple fundamental matrices to describe each moving object and segment the moving objects with epipolar geometry constraint. Unfortunately, this method is only valid for rigid objects and scenarios. For semantic video processing, some unsupervised methods adopted robust PCA method \cite{TPAMI2013_zhou,TPAMI2014_Gao,TIP2016_Li,AVSS2015_Sobral} for moving foreground estimation.
Later, long-term trajectory-based strategy \cite{ICCV2013_Papazoglou,CVPR2012_Ochs,CVPR2016_Xiao,BMVC2014_Faktor} becomes a common method in unsupervised VOS.
Depending on the analytic levels, the long-term trajectories are often generated on pixels \cite{CVPR2012_Ochs}, superpixels \cite{ICCV2013_Papazoglou,BMVC2014_Faktor,ECCV2014_Jain} or object proposals \cite{CVPR2016_Xiao,CVPR2015_Fragkiadaki,CVPR2017_Koh}, in which pixels with consistent trajectories are clustered as foreground and others are background. In order to obtain the accurate segmentation results, the long-term trajectory-based methods often take the entire video sequence as input, and thus they cannot segment moving objects in an online manner. In this paper, we mainly extract the moving objects by fusing the salient motion segmentation and object proposals.

\subsection{Motion Segmentation}
In an online VOS framework, motion segmentation between two adjacent frames is the key to segmenting moving objects frame-by-frame. 
Since the early geometry-based methods are sensitive to the selected model (\ie 2D homography or 3D fundamental matrix) \cite{Book2003_Hartley},
recent methods try to distinguish foreground/background with different motion cues.
Papazoglou and Ferrari \cite{ICCV2013_Papazoglou} first detected the motion boundaries based on the magnitude of optical flow field's gradient, 
and then used the filled binary motion boundaries to represent the moving regions.
However, this method is very sensitive to motion boundary extraction, ignoring object information.
In order to remove camera translation and rotation,
Bideau \etal \cite{ECCV2016_Bideau} utilized the angle and magnitude of optical flow to maximize the information about how objects are moving differently.
This method requires the focal length of camera to estimate its rotation and translation. However, given an arbitrary video sequence, the focal length of camera is often unknown.
Inspired by salient object detection methods on static images \cite{TPAMI2011_Cheng,ICCV2015_Zhang,CVPR2017_Hou,Info2018_Sun}, 
salient motion detection methods \cite{CVPR2016_Perazzi,TIP2013_Li} have been applied on optical flow field for moving object segmentation, where pixels with high motion contrast are classified as foreground. 
Due to the lack of object information, it cannot handle moving background (\eg \emph{moving water}) that do not indicate a generic object.

Recently, deep learning based methods have been widely applied in VOS. For example, Tokmakov \etal \cite{CVPR2017_Tokmakov} proposed an end-to-end CNN-based framework to automatically learn motion patterns from optical flow field, followed by an object proposals model and CRF model for segmentation refinement.
To fuse the motion and appearance information in a unified framework, Jain \etal \cite{CVPR2017_Jain} designed a two-stream CNN, where the appearance stream is used to detect object regions while the other motion stream is used to find moving regions.
In contrast to previous methods, we propose a new motion segmentation method with motion property in this paper. 
Specifically, since a segmented region should be \emph{moving} and indicate a \emph{generic object}, we apply off-the-shelf salient motion detection model \cite{ICCV2015_Zhang} and object proposal model \cite{ICCV2017_He} for accurate motion segmentation. Unlike other end-to-end deep learning based motion segmentation methods \cite{CVPR2017_Tokmakov, CVPR2017_Jain} that require a large number of training samples to learn motion patterns, our method directly uses a pretrained object proposal model without fine-tuning. 

\subsection{Semantic Segmentation with Object Proposals}
A comprehensive review on the topic of object proposal is out of the scope of this paper. Here, we only focus on the most related and recent works. The purpose of semantic segmentation \cite{ICCV2017_He,ECCV2014_Philipp,NIPS2015_Pinheiro} is to identify a set of generic objects in a given image with segmented regions. To generate object proposals, Kr{\"a}henb{\"u}hl \etal \cite{ECCV2014_Philipp} identified a set of automatically placed seed superpixels to hit all objects in the given image, then the foreground and background masks are generated by computing geodesic distance transform on these seeds. Finally, critical level sets is applied on the geodesic distance transform to discovering objects.
Recently, with the success of deep learning in object detection, 
DeepMask \cite{NIPS2015_Pinheiro} learns to propose object segment candidates with Fast R-CNN \cite{ICCV2015_Ross}.
He \etal \cite{ICCV2017_He} proposed Mask R-CNN framework for simultaneous instance-level recognition and segmentation.
By incorporating a mask branch for segmentation, Mask R-CNN extended Faster R-CNN \cite{NIPS2015_Ren} and achieved good segmentation results.
In this paper, we directly use a pretrained Mask R-CNN model to generate objectness map without fine-tuning.

\section{Our Approach}
\label{sec:proposed}

\subsection{Overview}
Let {\small $L^t_i \in \{0,1\}$} represent the foreground (denoted by $1$) or background (denoted by $0$) label of $i$-th pixel {\small $I^t_i$} in $t$-th video frame {\small $I^t$}.
Given an input video stream {\small $\Mat{I} = \{I^1,\ldots,I^T\}$}, 
our goal is to predict a set of binary foreground/background masks {\small $\Mat{L} = \{L^1, \ldots, L^T\}$} in a fully unsupervised and online manner.

In contrast to existing methods, our method is based on the motion property which requires the segmented region in VOS to be \emph{moving} and indicate a \emph{generic object}. We propose a new moving object segmentation framework by referring to the salient motion detection and object proposal methods. More specifically, for each frame\footnote{Notice that our method processes a given video in a frame-by-frame manner, which means the future frames are not processed.}, a salient motion detection method is applied to detect moving regions (\ie salient motion mask) and an object proposal method is used to detect generic objects (\ie objectness mask). Then, the detected results of this two methods are fused with our proposed fusion method (Section \ref{subsec_inter-frame}). The results by the salient motion detection and object proposal methods are not always reliable, especially for complex video scenes. To alleviate this problem, we propose a forward propagation refinement method to improve the segmentation results (Section \ref{subsec_multi-frame}). In addition, a CRF model is applied to further refine the results (Section \ref{subsec_refine}).

\subsection{Motion Segmentation}
\label{subsec_inter-frame}
In the following, salient motion mask represents moving regions while objectness mask denotes generic objects.
As mentioned, our motion segmentation is an effective fusion of salient motion segmentation and object proposal techniques. 
In the next section, we will introduce two techniques in sequence, followed by the proposed fusion method.

\subsubsection{Salient motion mask}
Motion reveals how foreground pixels move differently than their surrounding background ones. 
Thus, it is very useful for moving regions extraction. 
Unlike static camera environments studied in traditional background subtraction problems, 
foreground pixel displacements and camera movements are often unknown under freely moving camera environments. 

In this work, we employ saliency detection \cite{ICCV2015_Zhang} on optical flow to separate the moving regions from static background. 
This method computes global motion contrast of each pixel in a frame showing good performance for motion segmentation tasks \cite{CVPR2016_Perazzi,TIP2013_Li}.
Specifically, let {\small $\Mat{F}^t=\{F_1^t, F_2^t, \ldots, F_N^t\}$} be the backward optical flow field between two frames {\small $I^t$} and {\small $I^{t-1}$}, 
where each element {\small $F_i^t = [u_i^t, v_i^t]$} is the optical flow vector of pixel {\small $I_i^t$} in horizontal and vertical directions, 
$N$ is the total number of the frame pixels.
Let {\small $\tilde{S}^t$} be the salient motion map on optical flow field {\small $\Mat{F}^t$},
the global motion contrast {\small $\tilde{S}^t_i$} of each pixel {\small $I^t_i$} is computed as:
\begin{equation}
	\tilde{S}^t_i(\Mat{F}^t) = \sum_{\forall F_j^t \in \Mat{F}^t} d(F_i^t, F_j^t)
\label{eqn_sal_mot}
\end{equation}
where {\small $\tilde{S}^t_i \in [0, 1]$} and {\small $d(\cdot)$} is a distance metric~\cite{TPAMI2011_Cheng}.
For the sake of efficiency, we use the Minimum Barrier Distance (MBD) transform \cite{ICCV2015_Zhang} to detect salient motion.

\begin{figure}[!t]
	\centerline{\includegraphics[width=1.0\columnwidth]{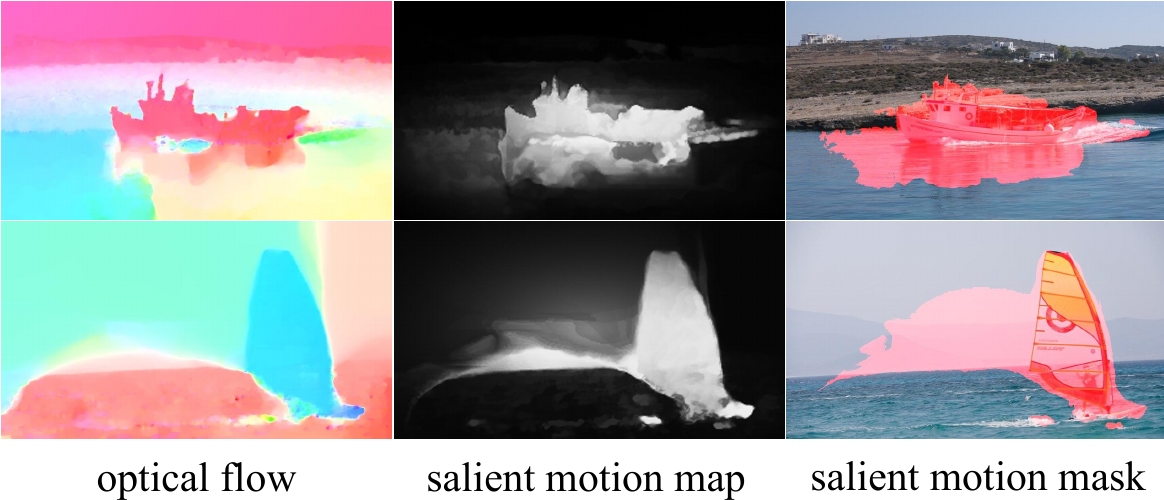}}
	\caption{Examples of inaccurate salient motion mask with moving background.}
	\label{fig_salmot}
\end{figure}

Given an unconstrained video sequence,
the object movements and camera motion are unknown.
In order to detect moving regions under various motion contrasts,
we utilize an adaptive threshold method \cite{TSMC1979_Otsu} to extract the salient motion map.
Then, pixels with high motion contrast are classified as foreground and the rest is background pixels. 
Let $\phi$ be the binary splitting function of our adaptive threshold method,
the salient motion mask {\small $S^t$} is computed as: 
\begin{equation}
S^t = \phi(\tilde{S}^t)
\label{eqn_sal_mask}
\end{equation}
where each element {\small $S_i^t \in \{0, 1\}$} denotes the binary foreground/background label of pixel {\small $I_i^t$}.

Different from moving object segmentation, salient motion mask only represents the moving regions.
Without any prior information about the object, moving background (\eg wave) may be classified as moving object (see \fig \ref{fig_salmot}).
Therefore, we incorporate object proposals to detect generic object.

\subsubsection{Objectness mask}  
As mentioned, salient motion segmentation method cannot differentiate moving objects from moving background.
Therefore, an object proposal technique is applied to extract generic objects. 
Based on the success of deep learning in object detection,
Mask R-CNN \cite{ICCV2017_He} extends Faster R-CNN algorithm \cite{NIPS2015_Ren} by adding a branch for predicting segmentation masks on each region of interest, 
and achieves the state-of-the-art detection and segmentation performance in static images. 
In this work, we use the pretrained Mask R-CNN \cite{ICCV2017_He} model in VOS to remove the moving background regions. 

\begin{figure}[!t]
	\centerline{\includegraphics[width=1.0\columnwidth]{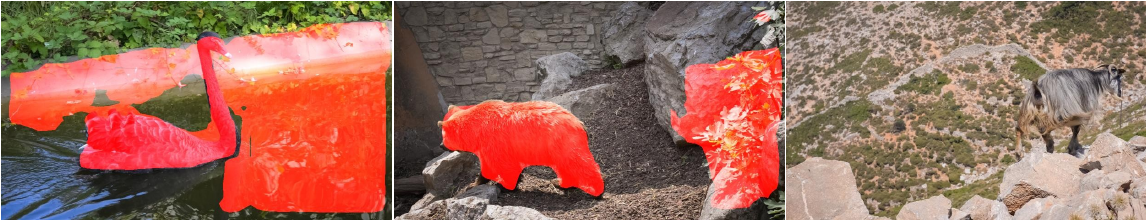}}
	\caption{Examples of unreliable objectness masks. 
		The first two images show that some background regions are wrongly classified as generic objects, 
		whereas the third image shows a missing object.}
	\label{fig_objness}
\end{figure}

In order to obtain an objectness mask {\small $O^t$} with high recall,
we set a low object confidence threshold (0.5 in our experiments) to extract the generic object regions.
Based on the binary objectness mask from Mask R-CNN, multiple segmented object regions can be obtained. 
Since the object region of interest also requires to satisfy the ``moving'' property, 
we directly use the binary objectness mask for fusion without any further processing, as illustrated in \fig \ref{fig_so}.
Though the object proposal model is not reliable enough in some complex video scenes, 
with false positive detections and missing objects shown in \fig \ref{fig_objness},
it still provides useful object information about the scenes.

It is worth mentioning that we directly use the Mask R-CNN model pretrained on MS-COCO dataset \cite{ECCV2014_Lin} without any further fine-tuning in our implementation.
In spite of that, it produces promising segmentation results on two benchmark datasets (see Section \ref{sec:exp-comp}). This demonstrates the potential of our method, since it is very different from many existing methods (such as \cite{CVPR2017_Khoreva,BMVC2017_Voigtlaender}) which require careful fine-tuning of the pretrained model for better results.
    
\subsubsection{Mask fusion}
As mentioned, the goal of motion segmentation is to detect moving objects. By computing the intersection region of the salient motion and the objectness mask, both moving and generic object properties can be satisfied. In the following, we describe our mask fusion method. 

\begin{figure}[t]
	\centerline{\includegraphics[width=1.0\columnwidth]{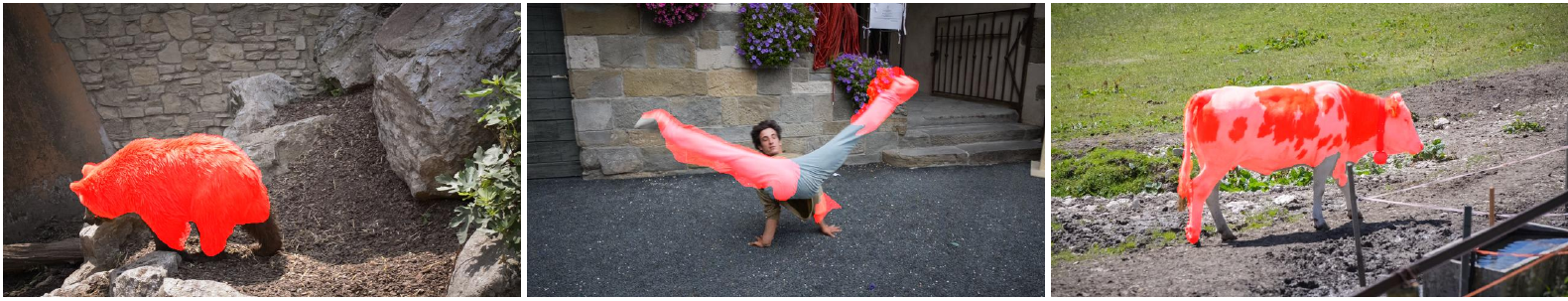}}
	\caption{Examples of incomplete salient motion segmentation when parts of the object move non-rigidly.
% 	Examples of incomplete salient motion segmentation when parts of the object move in non-rigid objects. 
	In the first image, the ``head'' and ``two legs'' of the bear are not completely moving; the second image demonstrates that only ``right leg'' and part of ``left leg'' of the man are moving; and the third image indicates that some ``legs'' remain stationary.}
	\label{fig_salmot-incomplete}
\end{figure}

In practice, directly extracting the intersection region may result in inaccurate segmentation. 
For example, as shown in \fig \ref{fig_salmot-incomplete}, 
when a part of the object moves in non-rigid objects, the segmentation results are incomplete to cover the whole object region.
To alleviate such problems, we first dilate the salient motion mask to produce moving regions with higher segmentation recall,
and then use the dilated moving regions for mask fusion.
Although some background regions may possibly be incorporated by the dilation operation,
our experiments show that it can be effectively removed by fusing it with the objectness mask.

Let {\small $S^t$} be the salient motion mask on optical flow field {\small $\Mat{F}^t$}, 
{\small $O^t$} be the objectness mask on current frame {\small $I^t$},
{\small $\mathcal{D}$} denote the image dilation function and $r$ represent the dilated radius. 
Then our fused segmentation mask {\small $P^t$} of frame {\small $I^t$} is computed by fusing the binary mask {\small $S^t$} and {\small $O^t$} as:
\begin{equation}
P^t = \mathcal{D}(S^t, r) \cap O^t
\label{eqn_so}
\end{equation}
where each element {\small $P_i^t \in \{0, 1\}$} denotes the binary foreground/background label of each pixel {\small $I_i^t$},
operator $\cap$ indicates the pixel-wise multiplication on the {\small $\mathcal{D}(S, r)^t$} and {\small $O^t$}.  
Our experiments on two benchmark datasets show that salient motion detection and object proposals are complementary to each other in VOS (see Section \ref{sec:exp-influence}).

\begin{figure}[!t]
	\centerline{\includegraphics[width=1.0\columnwidth]{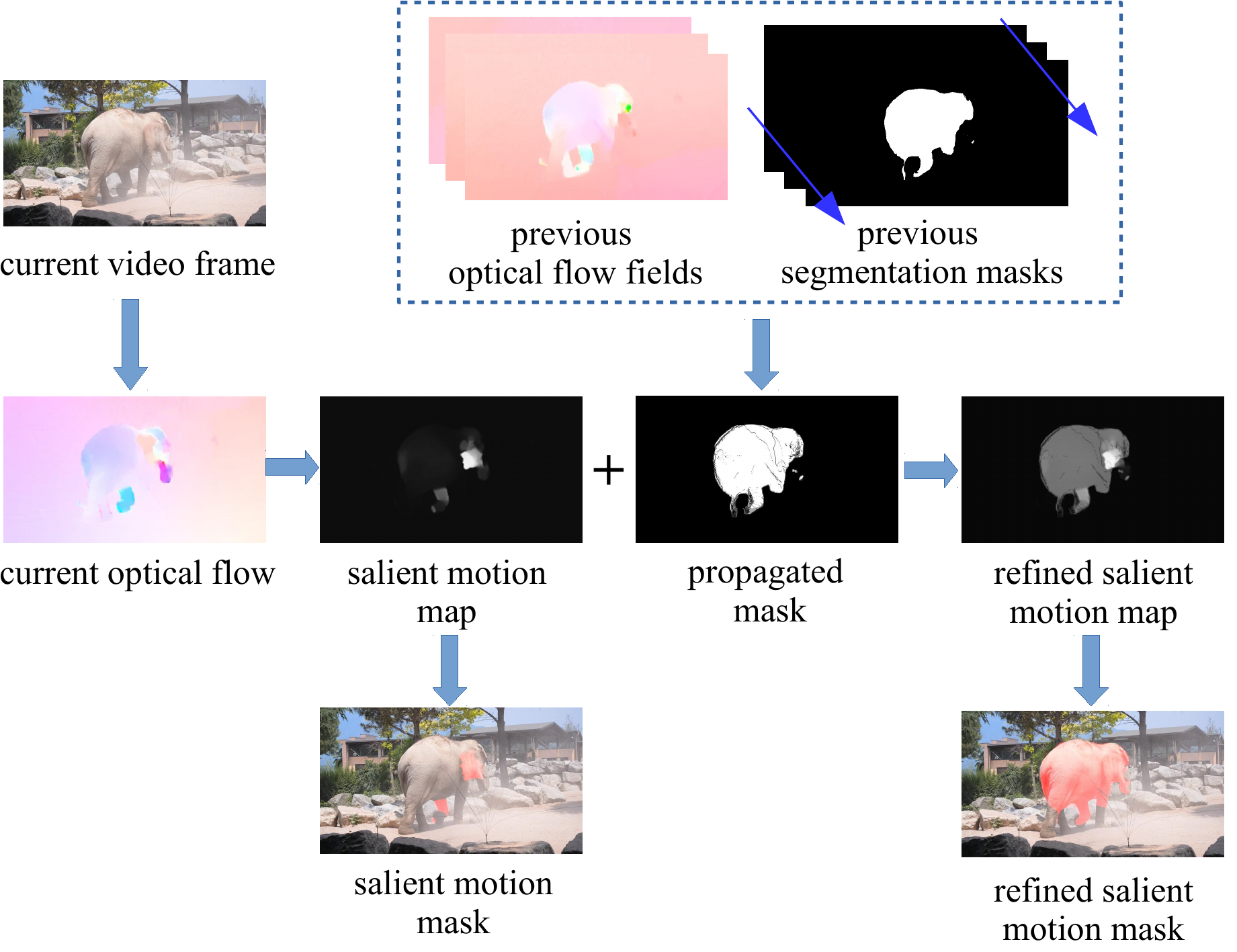}}
	\caption{An illustration of the forward propagation refinement. 
		The unreliable motion segmentation between two frames can be refined by propagating a set of previous segmentation masks.}
	\label{fig_multi-frame}
\end{figure}
  
\subsection{Forward Propagation Refinement}
\label{subsec_multi-frame}
In some complex video scenarios, it is difficult to obtain reliable salient motion detection and object proposals results on each frame 
(see \fig \ref{fig_salmot}, \ref{fig_objness} and \ref{fig_salmot-incomplete}).
Note that the video content in neighboring frames often share consistent motion dynamic. In other words, the content of the current frame is similar to the previous one. Therefore, we propose a forward propagation refinement method, which leverages the segmentation masks of previous frames for temporal mask consistency, and thus obtain more robust and accurate segmentation. 

Let {\small $P^t$} denote the segmentation mask of $t$-th frame without forward propagation refinement (namely, obtained by \eqn \ref{eqn_so}); {\small $M^t$} denotes the segmentation mask of $t$-th frame with the refinement method. For frame {\small $I^t$}, suppose we consider the segmentation masks of previous $n$ frames, \ie {\small $\{M^{t-n}, \ldots, M^{t-1}\}$}, which are propagated to the current frame (based on the pixel-wise tracking with optical flow) as {\small $\{\bar{M}^{t-n}, \ldots, \bar{M}^{t-1}\}$} for segmentation refinement. 

The refined salient motion map {\small $\bar{S}^t$} of current processing frame is recomputed with the original salient motion map {\small $\tilde{S}^t$} (obtained from \eqn \ref{eqn_sal_mot}) and propagated masks {\small $\{\bar{M}^{t-n}, \ldots, \bar{M}^{t-1}\}$} as:
\begin{equation}
\bar{S}^t = \theta \tilde{S}^t + (1-\theta) \sum_{\tau=1}^n \bar{M}^{t-\tau}
\label{eqn_sm}
\end{equation}
where {\small $\theta \in (0, 1)$} is a weight to balance these two components.
As shown in \fig \ref{fig_multi-frame}, the unreliable salient motion segmentation can be improved by forward propagating a set of previous segmentation masks. 

\begin{figure*}[!t]
	\centerline{\includegraphics[width=1.0\textwidth]{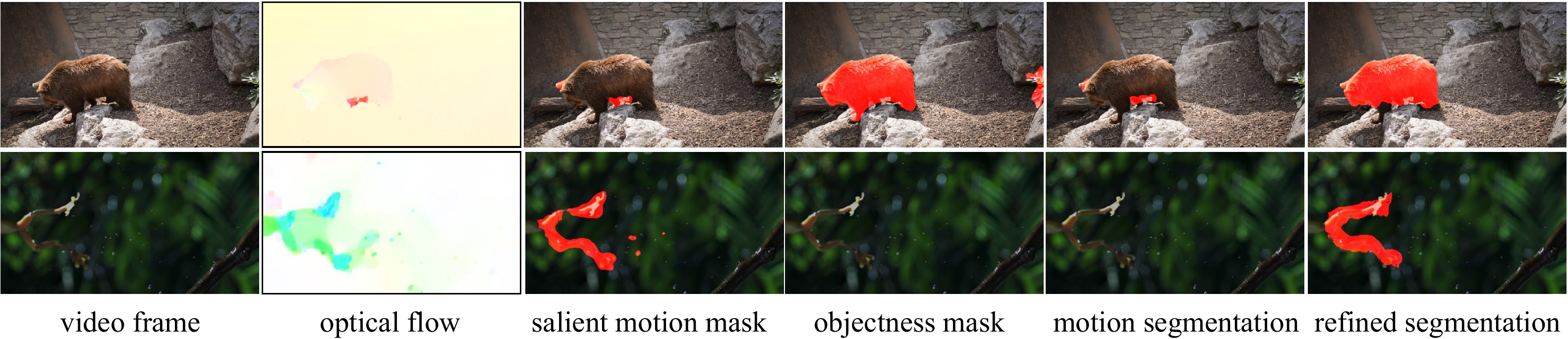}}
	\caption{Examples of the forward propagation refinement. The first row (video \emph{bear} in DAVIS-2016 dataset \cite{CVPR2016_Perazzi}) shows the improvement for inaccurate salient motion segmentation; and the second row (video \emph{frog} in SegTrack-v2 dataset \cite{ICCV2013_Li}) presents the improvement for unreliable object proposals, respectively.}
	\label{fig_som}
\end{figure*}

As the examples in \fig \ref{fig_so} show, given an arbitrary video sequence, the accuracy and robustness of both the motion and objectness components cannot be known in advance. Therefore, for the sake of simplicity, we use the same weight {\small $\theta$} to improve the objectness mask {\small $\bar{O}^t$} of current frame via:	
\begin{equation}
\bar{O}^t = \theta O^t + (1-\theta) \sum_{\tau=1}^n \bar{M}^{t-\tau}
\label{eqn_om}
\end{equation}

Similar to the motion segmentation between two frames, 
the improved segmentation mask {\small $M^t$} is obtained by fusing the refined masks {\small $\bar{S}^t$} and {\small $\bar{O}^t$} as:
\begin{equation}
M^t = \mathcal{D}(\phi(\bar{S}^t), r) \cap \phi(\bar{O}^t)
\label{eqn_som}
\end{equation}
where {\small $M^2=P^2$} indicates the initial motion segmentation between the first and second video frames.

Compared to the individually extracted motion segmentation between two frames,
by propagating previous segmentations to the current frame,
our method is able to improve both the unreliable salient motion segmentation and object proposals.
As shown in \fig \ref{fig_som}, based on the forward propagation refinement, the segmentation results are improved.

\subsection{CRF Refinement}
\label{subsec_refine}
Notice that the segmentation based on the motion cannot detect the object boundary very accurately in some cases \cite{ICCV2013_Papazoglou,CVPR2017_Tokmakov}. 
It may therefore degrade the results of our method even worse than the proposed forward propagation refinement (denoted by {\small $M^t$}). 
To alleviate this problem, the standard CRF model \cite{NIPS2011_Krahenbuhl} can be applied to our framework to further improving the final segmentation result, denoted as {\small $L^t$}. 
Based on the step-by-step processing strategy, we initialize the final segmentation label {\small $L^t$} with the binary mask {\small $M^t$}.

For segmentation optimization, we formulate our moving object segmentation task as a binary classification problem,
where the pixel labels are computed by constructing a graph \mbox{{\small $\mathcal{G}= \langle \mathcal{V}, \mathcal{E} \rangle$}}.
Here, {\small $\mathcal{V}$} denotes a set of vertices that correspond to the image pixels and {\small $\mathcal{E}$} represents edges that connect the four neighboring pixels.
The goal is to estimate optimal foreground/background label {\small $\mathcal{L}^t=\{L^t_1,L^t_2,...,L^t_N\}$} as:
\begin{equation}
\mathcal{L}^t = \arg\min_{L^t} E(L^t)
\label{eq_label}
\end{equation}
where {\small $L^t_i \in \{0,1\}$} is the label of each pixel and $0$ denotes background. 
The energy function for labeling {\small $L^t$} of all pixels is defined as:
\begin{equation}
E(L^t)=\sum_{i \in \mathcal{V}}\mathcal{U}^t_i(L^t_i)+\lambda \sum_{(i,j)\in \mathcal{E}}\mathcal{W}^t_{ij}(L^t_i,L^t_j)
\label{eq_graphcut}
\end{equation}
where {\small $\mathcal{U}^t_i(L^t_i)$} is the appearance based unary term.
{\small $W^t_{ij}(L^t_i,L^t_j)$} is the pairwise term for spatial smoothness purpose.
$\lambda > 0$ controls the relative effect of the two terms.

The unary term {\small $\mathcal{U}^t_i(L^t_i)$} models the deviations from the initially estimated foreground/background appearance in RGB color space.
Let {\small $\mathcal{C}^t_f$} be the total cost of assigning background to foreground and {\small $\mathcal{C}^t_b$} be the total cost of assigning foreground to background.
{\small $\mathcal{U}^t_i(L^t_i)$} is formulated as:
\begin{equation}
\mathcal{U}^t_i(L^t_i) = (1-L^t_i) \mathcal{C}^t_f+L^t_i \mathcal{C}^t_b
\end{equation}

\noindent
Taking account of the color Gaussian Mixture Model (GMM), the unary term is computed by a mixture of Gaussian probability distribution with $k_c=5$ components as in \cite{TOG2004_Rother}.

The pairwise term {\small $\mathcal{W}^t_{ij}(L^t_i,L^t_j)$} is used to ensure that neighboring pixels are assigned with the same label, 
which is computed by an exponential function as:
\begin{equation}
\mathcal{W}^t_{ij}(L^t_i,L^t_j)=(L^t_i-L^t_j)^2 exp(-\beta||I^t_i-I^t_j||^2) 
\end{equation}

\noindent
where {\small $\beta>0$} is a constant parameter, {\small $I^t_i$} and {\small $I^t_j$} are the intensity values of 4 neighboring pixels in frame {\small $I^t$}.
Then an efficient max-flow algorithm is applied to find the optimal labeling with minimal energy \cite{TOG2004_Rother}.

\begin{figure}[!t]
	\centerline{\includegraphics[width=0.5\textwidth]{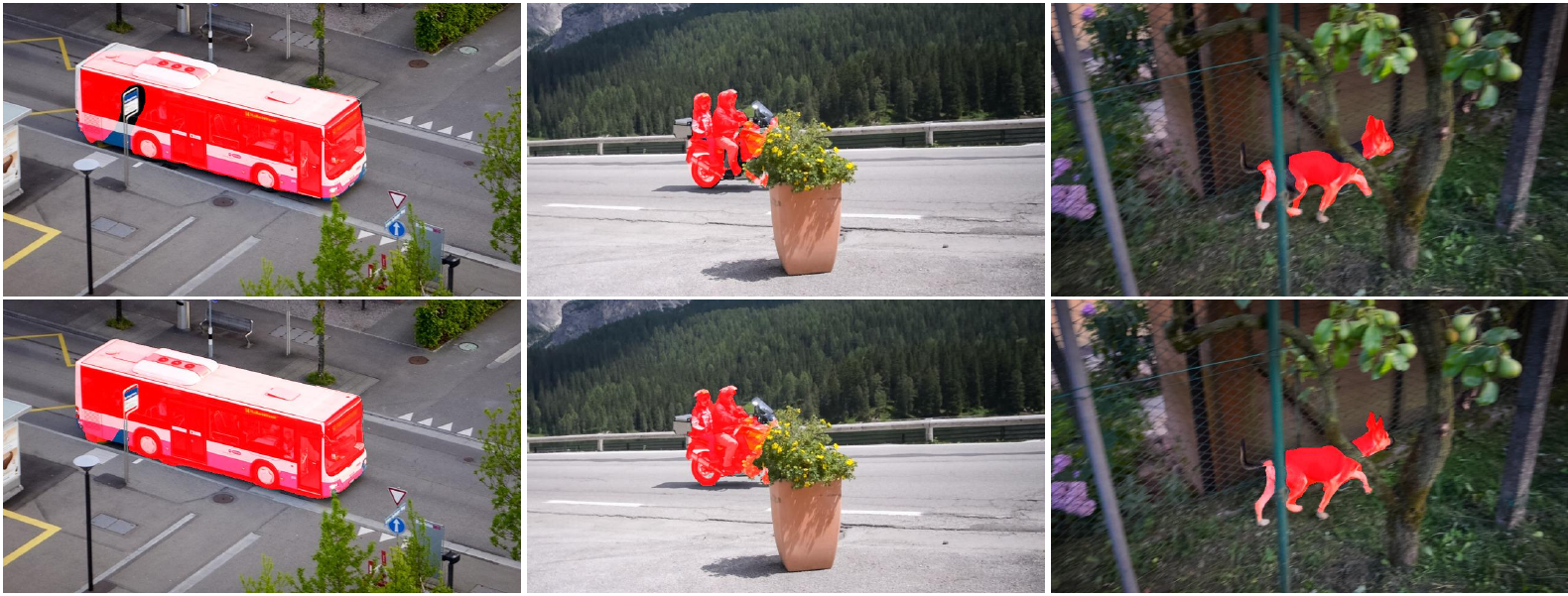}}
	\caption{Examples of segmentation refinement with CRF model. The first row shows the coarse segmentations whereas the second row shows the refined segmentations.}
	\label{fig_crf}
\end{figure}

The refined segmentations are shown in the second row of \fig \ref{fig_crf}, 
which improves the initial results (first row of \fig \ref{fig_crf}) on the object boundaries.
Finally, our entire approach is summarized in \alg \ref{alg_vos}. 

%The proposed motion segmentation is based on the salient motion detection and object proposals.
%However, when partial object moves, it may produce very high motion contrast while the rest object region shows low motion contrast.
%Therefore, when the motion contrast of the object region is not uniform,
%it is not reliable to apply the CRF refinement on the salient motion map.
%Different from the previous method \cite{CVPR2017_Tokmakov} that refining the segmentation on foreground probability map,
%we use the binary label value of multi-frame segmentation result $M^t$ to compute the unary terms and standard color-based pairwise terms.
%The refined segmentations are shown in the second row of \fig \ref{fig_crf}, 
%which improves over the initial segmentations (the first row of \fig \ref{fig_crf}) on the object boundaries.
%Finally, based on the step-by-step segmentation refinement, 
%the algorithm of our approach is described in \alg \ref{alg_vos}. 

\begin{algorithm}[!t]
	\small
	\caption{\small The proposed UOVOS framework}
	\label{alg_vos}
	\begin{flushleft}
		\textbf{Input:} Video stream {\small $\Mat{I} = \{I^1,\cdots,I^T\}$}, pretrained object proposals model $\mathcal{O}$,
		image dilation radius $r$, the number of used previous video frames $n$, accumulation weight $\theta$ \\
		\textbf{Output:} binary segmentation masks $\Mat{L} = \{L^1,\cdots,L^T\}$
	\end{flushleft}
	
	\SetAlgoLined
	\For{t = 2:T} {
		Optical flow field $F^t$ $\leftarrow$ $I^t$ and $I^{t-1}$, SIFT Flow \cite{TPAMI2011_Liu}    \\
		Salient motion map $\tilde{S}^t$ $\leftarrow$ $F^t$, MBD saliency \cite{ICCV2015_Zhang}   \\
		Salient motion mask $S^t$ $\leftarrow$ $\tilde{S}^t$, \eqn \ref{eqn_sal_mask}     \\
		Objectness mask $O^t$ $\leftarrow$ $\mathcal{O}$ and $I^t$, Mask R-CNN \cite{ICCV2017_He} \\
		Motion segmentation $P^t$ $\leftarrow$ $r$, $S^t$ and $O^t$, \eqn \ref{eqn_so}    \\
		\If{$t>n$} {
			Refined salient motion map $\bar{S}^t$ $\leftarrow$ $\theta$, $\tilde{S}^t$ and 
			propagated masks $\{\bar{M}^{t-n}, \cdots, \bar{M}^{t-1}\}$, \eqn \ref{eqn_sm} \\ 	
			
			Refined objectness map $\bar{O}^t$ $\leftarrow$ $\theta$, $O^t$ and 
			propagated masks $\{\bar{M}^{t-n}, \cdots, \bar{M}^{t-1}\}$, \eqn \ref{eqn_om} \\ 	
			
			Refined segmentation $M^t$ $\leftarrow$ $r$, $\bar{S}^t$ and $\bar{O}^t$, \eqn \ref{eqn_som}    \\
			
			CRF refinement $L^t$ $\leftarrow$ $M^t$ and $I^t$, CRF model \cite{NIPS2011_Krahenbuhl}  \\		
		}
	}
\end{algorithm}
\section{Experiments and Results}
\label{sec:experiment}
In this section, we first describe the implementation details, followed by the introduction of experimental datasets. Next, we detail the baselines and evaluation metrics, and finally report and analyze the experimental results. 

\subsection{Implementation Details}
\label{sec:exp-details}
Inspired by salient object detection on static images, previous works often applied salient object detection on optical flow field as salient motion detection, which has been demonstrated to be effective in \cite{CVPR2016_Perazzi,CVPR2017_Jain}. In this work, we adopt an efficient salient object detection method MBD \cite{ICCV2015_Zhang} on SIFT flow \cite{TPAMI2011_Liu} to detect the moving regions.
The objectness mask is detected by Mask R-CNN, which is the state-of-the-art method. In our implementation, we used the trained Mask R-CNN model (based on MS-COCO dataset) without any fine-tuning. We adopted the CRF model in \cite{NIPS2011_Krahenbuhl} for final segmentation refinement.
It is worth mentioning that, for all the above models, we used the provided default parameters of these approaches without any fine-tuning, and we achieve the state-of-the-art performance ( as shown in Section \ref{sec:exp-comp}). 

Without additional specification hereafter, the reported results are based on the following parameter settings: for object proposals, the confidence threshold of object detection is set to $0.5$ and the radius for image dilation operator is $6$.
The Otsu's method \cite{TSMC1979_Otsu} is used for adaptive threshold segmentation. The number of adaptive thresholds is set to 3 for salient motion segmentation and 2 for multi-frame object mask in our experiments.
Besides, the number of previous frames for forward propagation refinement method is set to 2 (\ie $n = 2$ in Section \ref{subsec_multi-frame}).

Our method is mainly implemented in MATLAB and evaluated on a desktop with 1.7GHz Intel Xeon CPU and 32GB RAM.
Given an image of resolution $480 \times 854$ pixels,
the average processing time of the key components is shown in Table \ref{tbl_time}. 
From the table, we can see that the main computational cost of our approach lies in the optical flow estimation component, while the other components are very fast.

\begin{table}
	\centering 
    \caption{The run time of each component.}
    \begin{tabular}{L{32ex}|C{12ex}} 
    \toprule
    Component                       & Runtime (s) \\ \midrule
    Optical flow                    & 8.0~~~      \\
    Salient motion detection        & 0.01        \\
    Object proposals                & 0.3~~~      \\
    Forward propagation refinement  & 0.05        \\ 
    CRF refinement                  & 1.6~~~      \\ 
    \bottomrule
    \end{tabular}
    \label{tbl_time}
\end{table}
 
\begin{figure*}[!t]
	\centerline{\includegraphics[width=1.0\textwidth]{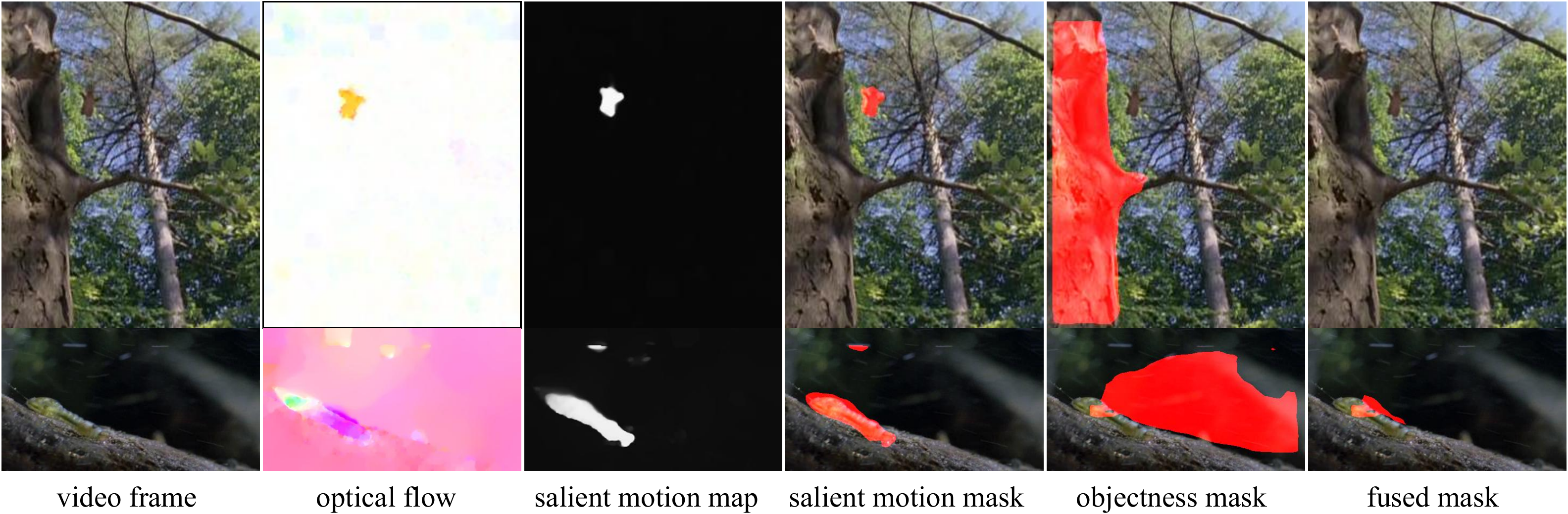}}
	\caption{Examples of failed mask fusion on SegTrack-v2 dataset~\cite{ICCV2013_Li}. 
	Due to the low video resolution and cluttered background, 
	Mask-RCNN \cite{ICCV2017_He} failed to detect accurate object proposals with pretrained model. 
	The video in first row and second row is \emph{birdfall} and \emph{worm}, respectively.}
	\label{fig_fail-segtrack}
\end{figure*}

\subsection{Datasets and Evaluation Metrics}
To test the performance of our method, 
we evaluate it on two densely annotated (ground-truth masks on all video frames) VOS benchmark datasets: a high resolution DAVIS-2016 dataset~\cite{CVPR2016_Perazzi} and a low resolution SegTrack-v2 dataset~\cite{ICCV2013_Li}. Beside, we report the performance of the proposed method on a sparsely annotated (ground-truth masks on a few video frames only) benchmark dataset FBMS-59 \cite{TPAMI2014_Ochs}.

\subsubsection{Datasets}
The DAVIS-2016 dataset~\cite{CVPR2016_Perazzi} is currently the most challenging VOS benchmark, 
which contains 50 high resolution video sequences of diverse object categories and 3455 densely annotated pixel-wise ground-truth. 
Videos in this dataset are unconstrained and the challenging problems include appearance change, dynamic background, fast-motion, motion blur and occlusion.

SegTrack-v2 dataset \cite{ICCV2013_Li} is a widely used benchmark for VOS,
which consists of 14 low resolution videos with a total of 1066 frames. The ground-truth of this dataset is also pixel-wise annotated. 
The main challenges in SegTrack-v2 dataset include drastic appearance change, complex background, occlusion, abrupt motion and multiple moving objects.
Similar to previous methods \cite{BMVC2014_Faktor,CVPR2017_Jain},
we treated multiple objects with individual ground-truth as a single foreground for evaluation.

FBMS-59 dataset is composed of 59 videos, in which 29 are used for training and 30 for evaluation. Similar to the previous work \cite{CVPR2017_Koh}, we report the performance of our method on 30 test videos for comparison. Besides, since the FBMS-59 dataset contains multiple moving objects, we also convert them to a single foreground.

\subsubsection{Evaluation metrics}
For quantitative analysis, 
the standard evaluation metrics: region similarity $\mathcal{J}$, contour accuracy $\mathcal{F}$ and temporal stability $\mathcal{T}$ are adopted. 
Region similarity $\mathcal{J}$ is defined as the mean \emph{Intersection-over-Union} (mIoU) of the estimated segmentation and the ground-truth mask.
$\mathcal{F}$ measures the accuracy of the contours and $\mathcal{T}$ measures the temporal stability of the segmentation results in VOS. 
More description about the evaluation metrics can be found in \cite{CVPR2016_Perazzi}.
For performance comparison between the proposed segmentation and the state-of-the-art approaches, 
we utilized the provided codes and parameter configurations from the benchmark website\footnote{\url{https://graphics.ethz.ch/~perazzif/davis/code.html}}.
Since mIoU denotes the region similarity between the segmentation result and ground-truth, 
we mainly analyze the performance of each algorithm with mIoU metric as in previous works \cite{CVPR2017_Tokmakov,BMVC2014_Faktor,CVPR2017_Koh,CVPR2017_Jain}.

\begin{table}
	\centering 
	\caption{Ablation study of our method with mIoU metric.
		The improvement of $\mathcal{S}$+$\mathcal{O}$ is compared to the salient motion segmentation $\mathcal{S}$.}
    \begin{tabular}{L{18ex}|C{15ex}|C{15ex}} \toprule
    Fused components                                          & DAVIS-2016         & SegTrack-v2 \\ \midrule
    $\mathcal{S}$                                             & 57.1               & 47.3        \\
    $\mathcal{O}$                                             & 57.1               & 54.0        \\
    $\mathcal{S}$ + $\mathcal{O}$                               & 69.6 (+12.5)       & 55.3 (+8.0)        \\
    $\mathcal{S}$ + $\mathcal{O}$ + $\mathcal{P}$                 & 74.6 (~~+5.0)        & 61.5 (+6.2)        \\
    $\mathcal{S}$ + $\mathcal{O}$ + $\mathcal{P}$ + $\mathcal{C}$   & {\bf 77.2} (~~+2.6)  & {\bf 64.3} (+2.8)  \\ \bottomrule
    \end{tabular}
    \label{tbl_modality}
\end{table}

\subsection{Ablation Studies}
\label{sec:exp-influence}
To demonstrate the influence of each component in the proposed method,
we reported the performance of different modalities fusion on two densely annotated datasets DAVIS-2016 \cite{CVPR2016_Perazzi} and SegTrack-v2 \cite{ICCV2013_Li}. 

To demonstrate the robustness and effectiveness of each component, we set $\theta=0.9$ and $n=2$. Besides, \emph{all parameters in our method are kept same on these two datasets for performance evaluation.} 
For ease of presentation, we denote the key component of our approach as follows.
\begin{itemize}
	\item $\mathcal{S}$: salient motion segmentation on optical flow field.
	\item $\mathcal{O}$: object proposals on current video frame.
	\item $\mathcal{P}$: forward propagation refinement with several previous segmentations.
	\item $\mathcal{C}$: coarse-to-fine segmentation with CRF.
\end{itemize}
Based on these components,
the improvements of each additional component are reported in \tab \ref{tbl_modality}.
Next, we detailedly analyze the effectiveness of each component in our approach.

\subsubsection{Effectiveness of the mask fusion.}
As a reminder, the mask fusion is to remove some potential segmentation noise, such as moving background and static objects. The moving regions and object regions are detected by the salient motion detection method and object proposal method, respectively.  

As shown in \tab \ref{tbl_modality},
on DAVIS-2016 dataset, 
the mIoU of salient motion detection $\mathcal{S}$ is $57.1\%$, which denotes the accuracy of moving region segmentation.
Similarly, the performance of object proposals $\mathcal{O}$ is $57.1\%$, which denotes the accuracy of object region segmentation.
Based on our pixel-wise fusion method,
moving background regions and static objects can be effectively removed.
Compared to the salient motion detection component $\mathcal{S}$, 
the mIoU after fusion ($\mathcal{S}$+$\mathcal{O}$, 69.6\%) is significantly improved by an absolute gain of $12.5\%$.

Similarly, on SegTrack-v2 dataset,
the mIoU of salient motion detection component $\mathcal{S}$ is $47.3\%$ and the object proposals $\mathcal{O}$ is $54.0\%$.
Based on the proposed mask fusion method, the fused results $\mathcal{S}$+$\mathcal{O}$ ($55.3\%$) have achieved an absolute gain of $8.0\%$ compared to $\mathcal{S}$ ($47.3\%$).

Because the videos in SegTrack-v2 dataset are of low-resolution, 
the semantic object segmentation results of the object proposals model pretrained on MS-COCO dataset \cite{ECCV2014_Lin} are not very good on some videos, and thus the improvement is not as high as in DAVIS-2016 dataset.
As show in \fig~\ref{fig_fail-segtrack}, the moving objects are accurately extracted by salient motion segmentation.
However, due to low video resolution and cluttered background in some complex scenes,
Mask-RCNN \cite{ICCV2017_He} failed to provide accurate generic object regions with the direct use of pretrained model.
Therefore, it is expected that the performance of our method can be further improved by fine-tuning the object proposal model. 

\subsubsection{Effectiveness of the forward propagation refinement}
In order to handle the unreliable salient motion detection and object proposals in individual video frame,
we propose a forward propagation refinement method to improve the segmentation accuracy. 
As shown in \tab \ref{tbl_modality},
compared to the motion segmentation $\mathcal{S}$+$\mathcal{O}$ between two video frames,
the forward propagation refinement ($\mathcal{S}$+$\mathcal{O}$+$\mathcal{P}$) can achieve absolute gain of $5.0\%$ and $6.2\%$ on DAVIS-2016 dataset and SegTrack-v2 dataset, respectively. 
Although the object movements and video quality are very different in these two datasets, 
the proposed method is still robust for both conditions.

\subsubsection{Effectiveness of the CRF refinement}
We also applied a CRF model for result refinement, and the segmentation accuracy can be further improved.
As shown in \tab \ref{tbl_modality}, we achieve absolute gain of $2.6\%$ and $2.8\%$ on DAVIS-2016 dataset and SegTrack-v2 dataset, respectively.
From the above results and analysis, we can see that each component of our model is very useful and can indeed improve the performance.

\subsection{Influence of Key Parameters}
\label{sec:exp-param}
In this section, we analyze the influence of key parameters in our approach, including the accumulation weight $\theta$ and frame number of $n$ for forward propagation refinement on two densely annotated DAVIS-2016 and SegTrack-v2 datasets. 

\subsubsection{Weight $\theta$}
$\theta$ is the weight that decide the contribution of the previous frames' segmentation results affect the current frame segmentation. 
When $\theta=1.0$, it denotes that no information from the previous frame propagates to the current one. As shown in the left of \fig \ref{fig_parameters}, when the value of $\theta$ decreases from $1.0$ to $0.7$, the performance increases first and then slightly decreases on both datasets. The best performance is achieved by $\theta=$ 0.85 and 0.75 for DAVIS-2016 dataset and SegTrack-v2 dataset, respectively. Notice that the smaller the value of $\theta$, the more information (segmentation results) from the previous frames are propagated to the current frame. Therefore, when $\theta$ becomes too small, the information from previous frames becomes dominating and thus deteriorates the performance\footnote{The extreme case is when $\theta$ set to 0, which denotes that the information from previous frames overwrite the current frame and mislead the result.}. In our experiments, when $\theta$ is reduced to 0.7, the performance on both datasets is still better than $\theta=1.0$. This demonstrates that the proposed component (\ie forward propagation refinement) is quite robust and can improve the performance within a wide range of $\theta$.

\subsubsection{Frame number $n$}
\begin{figure}[!t]
	\includegraphics[width = 0.23\textwidth, height = 0.15\textheight]{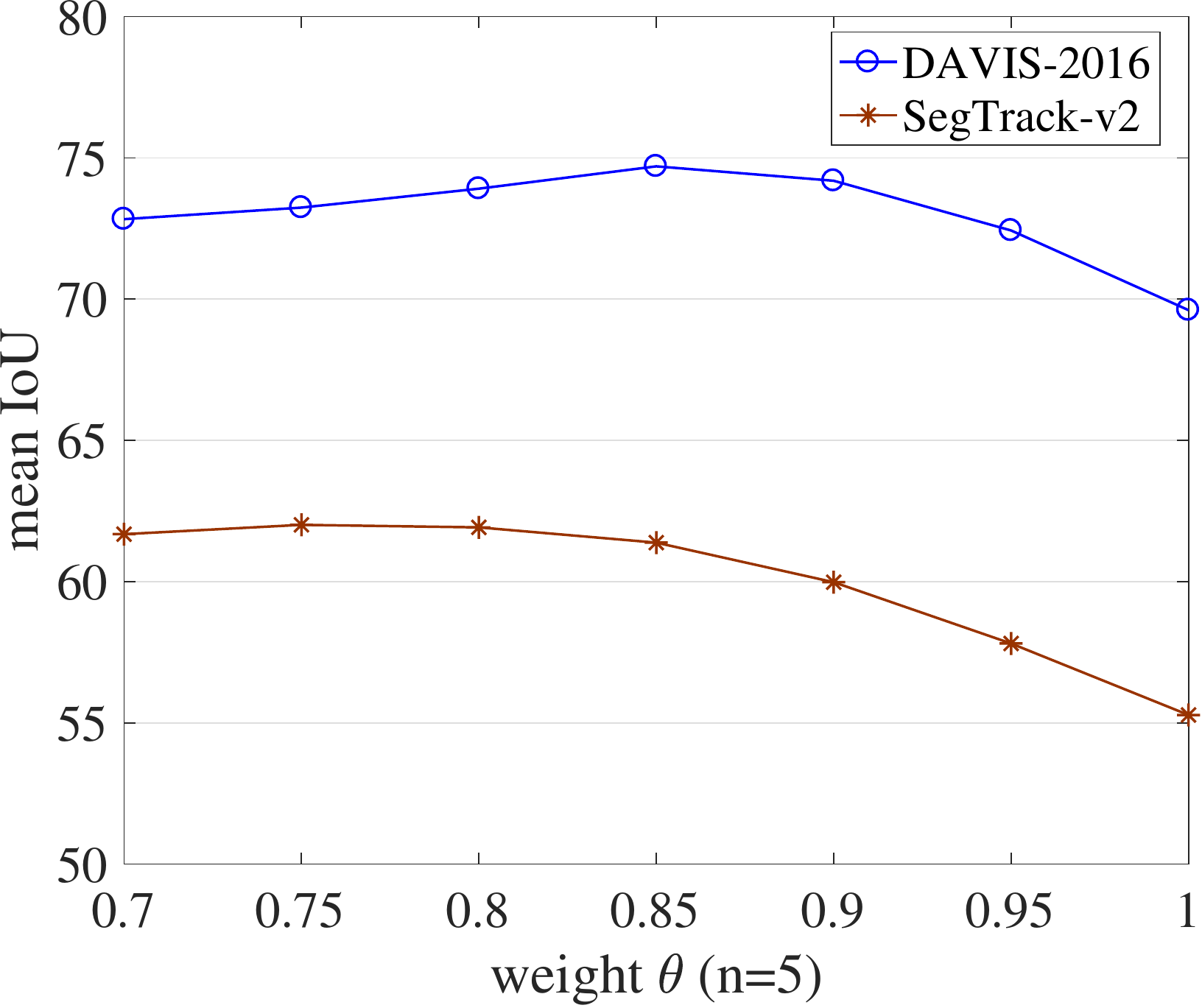}
	\quad
	\includegraphics[width = 0.23\textwidth, height = 0.15\textheight]{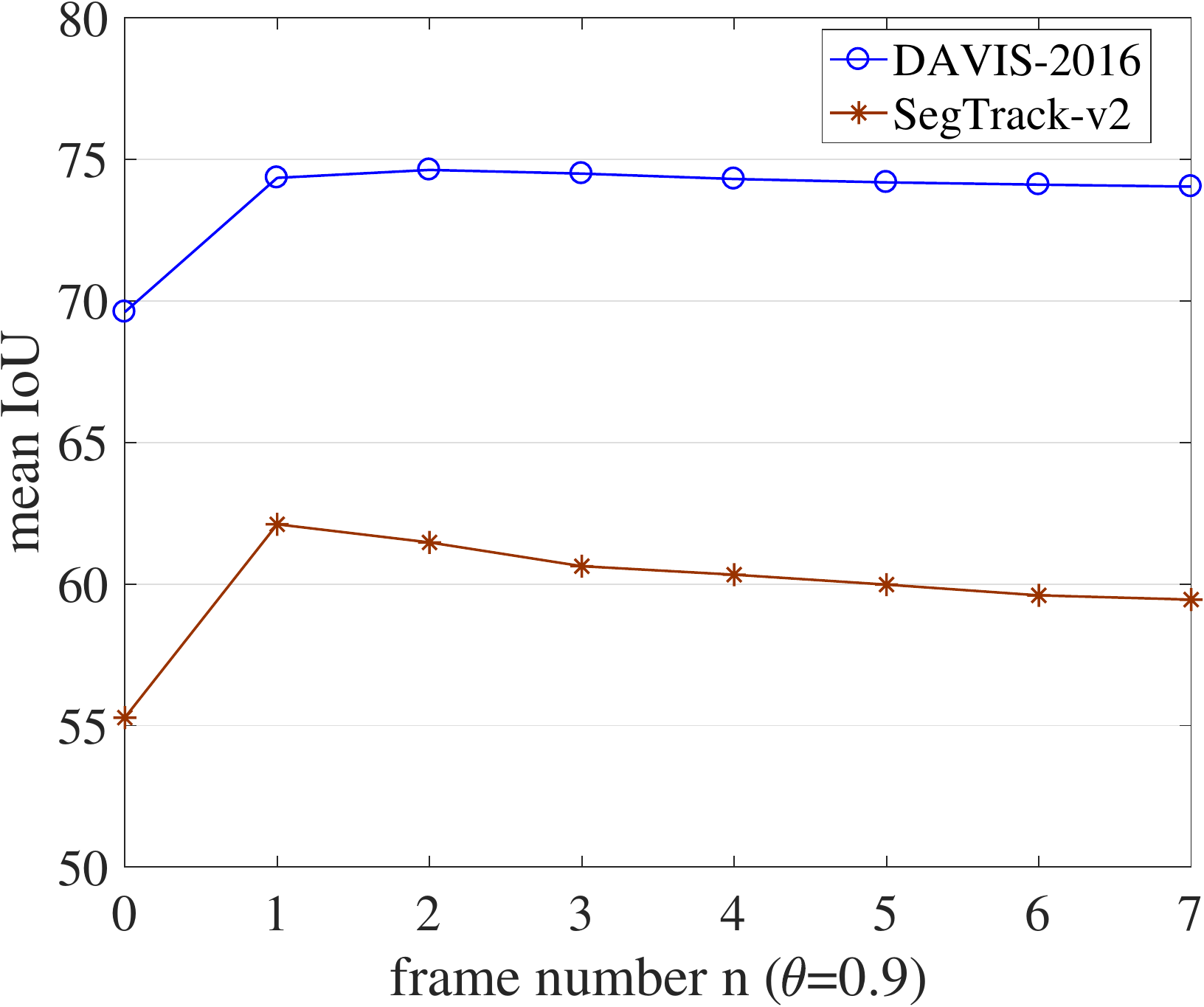}
	\caption{Performance analysis with different weight $\theta$ and frame number $n$ for forward propagation refinement on DAVIS-2016 and SegTrack-v2 dataset. $\theta=1$ and $n=0$ denote the motion segmentation between two video frames, respectively. It can be seen that the proposed forward propagation refinement can improve the accuracy within a wide range of the parameters. Thus, it is not sensitive to these two parameters in both two datasets.}
	\label{fig_parameters}
\end{figure}

Another key parameter is the number of previous frames, 
which decides how many previous segmentation masks are used for forward propagation refinement.
$n=0$ denotes the motion segmentation between two adjacent video frames.
We analyze the forward propagation refinement with various $n$ values and $\theta$ set to 0.9.
the performance is shown in the right of \fig \ref{fig_parameters}.
From the results, we can see that the performance can be improved when $n \in [1,7]$ on both datasets. 
The larger the n is, the more previous frames are considered. 
When n is very large (\eg $n = 100$), it means the information of frames which are far from the current frame (\ie the 100-th frame before this frame) is also considered, which may lead to noisy information\footnote{It is highly possible to introduce noisy information as the frame which is far from the current frame may contain very different content.}.
In particular, the proposed method achieves the best performance of $74.63\%$ when $n=2$ on DAVIS-2016 dataset and $62.11\%$ when $n=1$ on SegTrack-v2 dataset.
Because SegTrack-v2 dataset is having lower image resolution, the object proposals in SegTrack-v2 dataset is not as reliable as in DAVIS-2016 dataset, 
which is why the performance variation in DAVIS-2016 dataset is smoother, as shown in \fig \ref{fig_parameters}.

\subsection{Comparison to the State-of-the-art Methods}
\label{sec:exp-comp}

\subsubsection{Baselines}
We compared our method with several state-of-the-art unsupervised moving object segmentation methods to verify the effectiveness of our method. Based on whether they operate in offline or online manner, we group these competitors into two categories.

{\bf Unsupervised offline methods:} 
To achieve good segmentation performance, offline methods often require the entire video sequence to generate long-term trajectories,
and the moving objects are identified by motion or objectness cues. 
Based on the provided results of DAVIS-2016 dataset,
the compared baselines include: ARP \cite{CVPR2017_Koh}, FST \cite{ICCV2013_Papazoglou}, NLC \cite{BMVC2014_Faktor}, MSG \cite{ICCV2011_Ochs}, KEY \cite{ICCV2011_Lee} and TRC \cite{CVPR2012_Fragkiadaki}, STP~\cite{CVPR2013_Zhang} and ACO~\cite{CVPR2016_Jang}.

{\bf Unsupervised online methods:} 
Instead of generating long-term trajectories on the entire video sequence, online methods are able to segment the moving objects in a frame-by-frame manner. 
The compared baselines include: FSEG \cite{CVPR2017_Jain}, LMP \cite{CVPR2017_Tokmakov}, CVOS \cite{CVPR2015_Taylor}, SAL \cite{CVPR2015_Wang} and SFM \cite{CVPR2012_Perazzi}.
To be specific, 
FSEG \cite{CVPR2017_Jain} and LMP \cite{CVPR2017_Tokmakov} are deep learning based methods which attempt to learn the moving patterns from optical flow field.
FSEG \cite{CVPR2017_Jain} fuses the appearance and motion in a two-stream fully convolutional neural network,
where the appearance-stream is used to extract the candidate object regions while the motion-stream is used to produce the moving foreground.
LMP \cite{CVPR2017_Tokmakov} is also a fully convolutional network, which is learned from synthetic videos with ground-truth optical flow and motion segmentation. Based on the coarse motion segmentation, LMP adopts object proposals and CRF to refine the initial result.
CVOS \cite{CVPR2015_Taylor} automatically segments moving objects with several frames, then a tracking strategy is used to propagate the initialized mask to subsequent frames.
SAL \cite{CVPR2015_Wang} is based on spatio-temporal saliency detection and performs VOS on multiple frames for online processing.
SFM \cite{CVPR2012_Perazzi} is a salient motion detection method that operates between two adjacent frames.

% \usepackage{multirow}
\begin{table*}[!t]
	\centering
	\caption{Overall results of {\it region similarity} ($\mathcal{J}$), {\it contour accuracy} ($\mathcal{F}$) and {\it temporal stability} ($\mathcal{T}$) on DAVIS-2016 TrainVal dataset. The best results are marked in {\bf Bold Font}. Our method achieves significant improvements on the mean and recall with both $\mathcal{J}$ and $\mathcal{F}$ metrics. Besides, the proposed method outperforms the unsupervised offline approach ARP~\cite{CVPR2017_Koh} on all metrics.}
	\label{tbl_overall-davis}
	\resizebox{\textwidth}{!}{
		\begin{tabular}{c|c|c|c|c|c|c|c||c|c|c|c|c|c} \toprule
			\multicolumn{2}{c|}{\multirow{3}{*}{Measure}}  & \multicolumn{6}{c||}{Offline} & \multicolumn{6}{c}{Online}                                                               \\ \cmidrule{3-14} 
			\multicolumn{2}{c|}{} & ARP~\cite{CVPR2017_Koh} & FST~\cite{ICCV2013_Papazoglou} & NLC~\cite{BMVC2014_Faktor} & MSG~\cite{ICCV2011_Ochs} & KEY~\cite{ICCV2011_Lee} & TRC~\cite{CVPR2012_Fragkiadaki} & FSEG~\cite{CVPR2017_Jain} & LMP~\cite{CVPR2017_Tokmakov} & CVOS~\cite{CVPR2015_Taylor} & SAL~\cite{CVPR2015_Wang} & SFM~\cite{CVPR2012_Perazzi} & UOVOS                         \\ \midrule \midrule
			\multirow{3}{*}{$\mathcal{J}$} & Mean $\uparrow$    & {\bf 76.3} & 57.5 & 64.1 & 54.3      & 56.9       & 50.1 & 71.6      & 69.7 & 51.4       & 42.6 & 53.2 & {\bf 77.8} \\  
			                               & Recall $\uparrow$  & {\bf 89.2} & 65.2 & 73.1 & 63.6      & 67.1       & 56.0 & 87.7      & 82.9 & 58.1       & 38.6 & 67.2 & {\bf 93.6} \\ 
			                               & Decay $\downarrow$ & ~~3.6      & ~~4.4  & ~~8.6  & ~~{\bf 2.8} & ~~7.5        ~~& 5.0  & ~~{\bf 1.7} & ~~5.6  & 12.7       & ~~8.4  & ~~5.0  & ~~2.1        \\ \midrule
			\multirow{3}{*}{$\mathcal{F}$} & Mean  $\uparrow$   & {\bf 71.1} & 53.6 & 59.3 & 52.5      & 50.3       & 47.8 & 65.8      & 66.3 & 49.0       & 38.3 & 45.2 & {\bf 72.0} \\ 
			                               & Recall $\uparrow$  & {\bf 82.8} & 57.9 & 65.8 & 61.3      & 53.4       & 51.9 & 79.0      & 78.3 & 57.8       & 26.4 & 44   & {\bf 87.7} \\  
			                               & Decay $\downarrow$ & ~~7.3      & ~~6.5  & ~~8.6  & ~~5.7       & ~~7.9        & ~~6.6  & ~~{\bf 4.3} & ~~6.7  & 13.8       & ~~7.2  & ~~5.3  & ~~3.8        \\ \midrule
			$\mathcal{T}$                  & Mean $\downarrow$  & 35.9       & 29.3 & 36.6 & 26.3      & {\bf 21.0} & 34.5 & 29.5      & 68.8 & {\bf 25.6} & 60.0 & 65.0 & 33.0       \\ \bottomrule
		\end{tabular}
	}
\end{table*}

\subsubsection{Quantitative Analysis}
To demonstrate the performance of our approach, 
we compare it with several unsupervised methods on DAVIS-2016 \cite{CVPR2016_Perazzi} dataset and SegTrack-v2 \cite{ICCV2013_Li} dataset.
The quantitative comparison results on DAVIS-2016 dataset and SegTrack-v2 dataset are shown in \tab \ref{tbl_overall-davis} and \tab \ref{tbl_video-segtrack}. 
In addition, the compared algorithms and results on SegTrack-v2 dataset are obtained from a recent work \cite{CVPR2017_Jain}.
Similar to \cite{CVPR2017_Jain}, we mainly analyze our method on the larger DAVIS-2016 dataset.

{\bf Performance on DAVIS-2016:}
Based on the optimum parameters from Section~\ref{sec:exp-param}, we report the UOVOS results with parameter setting: $\theta=0.85$ and $n=5$ on DAVIS-2016 dataset. \tab \ref{tbl_overall-davis} shows the performance of our method with the region similarity $\mathcal{J}$ and contour accuracy $\mathcal{F}$. It can be seen that our method achieves the best performance among all of the compared algorithms, including the best offline method ARP.
Especially, our approach obtains significant improvement in recall of both region similarity $\mathcal{J}$ (93.6\%) and contour accuracy $\mathcal{F}$ (87.7\%), which can achieve absolute gain of $4.4\%$ and $4.9\%$ respectively when compared to the best offline method ARP.
Moreover, the decay of $\mathcal{J}$ and $\mathcal{F}$, and temporal stability $\mathcal{T}$ of our method are also better than ARP.

Because our method is an online one, we mainly analyze the comparisons with the state-of-the-art online methods.
FSEG and LMP adopt an end-to-end deep learning framework for motion segmentation between two adjacent frames,
and both of them fuse the optical flow field and object proposals for moving object segmentation.
In contrast, our method is based on salient motion detection and object proposals, and thus it does not require further training on a large number of well-annotated data. Besides, since the video content often continuously changes, we use the important temporal connection of the video content for mask propagation among frames. As shown in \tab \ref{tbl_overall-davis}, our method significantly outperforms the compared ones by a large margin.
Specifically, our method outperforms FSEG by $6.2\%$ and LMP by $8.1\%$ on $\mathcal{J}$ metric. The online method CVOS is very sensitive to the object initialization and it suffers the drift problem when tracking the initialized object mask. As shown in \tab \ref{tbl_overall-davis}, due to the unreliable online segmentation strategy, the accuracy of CVOS is only $51.4\%$.
Another online approach SAL uses spatio-temporal saliency detection method to extract moving object regions. 
However, as the moving object is not always salient in some videos, and thus their segmentation result (42.6\%) is also not good enough.
SFM is a salient motion detection method, because it has not considered the object information and temporal connection of the video content, its segmentation result (53.2\%) is also not very good.

\begin{table}[!t]
	\centering
	\caption
		{Video object segmentation results on Segtrack-v2 dataset with mIoU metric.
		As reported in FSEG \cite{CVPR2017_Jain}, we show the results of all 14 videos.	The results of NLC are the mIoU over 12 videos as in their paper \cite{BMVC2014_Faktor}. Our method outperforms several state-of-art methods, which include the two-stream deep learning based approach FSEG \cite{CVPR2017_Jain}.
		}
	\label{tbl_video-segtrack}
	\resizebox{1.0\columnwidth}{!}{
	\begin{tabular}{c|C{10ex}|C{10ex}|C{10ex}||C{10ex}|C{10ex}}
		\toprule
		\multirow{2}{*}{Video} & \multicolumn{3}{c||}{Offline}                                                          & \multicolumn{2}{c}{Online}              \\ \cmidrule{2-6} 	
		                       & FST~\cite{ICCV2013_Papazoglou} & KEY~\cite{ICCV2011_Lee} &  NLC~\cite{BMVC2014_Faktor} & FSEG \cite{CVPR2017_Jain}  & UOVOS      \\ \midrule \midrule
		birdfall               & 17.5                           & 49.0                    & {\bf 74.0}                  & {\bf 38.0}                 & 13.9       \\  
	  bird\_of\_paradise     & 81.8                           & {\bf 92.2}              & -                           & 69.9                       & {\bf 79.7} \\
		bmx                    & 67.0                           & 63.0                    & {\bf 79.0}                  & 59.1                       & {\bf 62.4} \\
		cheetah                & 28.0                           & 28.1                    & {\bf 69.0}                  & {\bf 59.6}                 & 56.5       \\
		drift                  & 60.5                           & 46.9                    & {\bf 86.0}                  & {\bf 87.6}                 & 84.3       \\
		frog                   & 54.1                           & 0.0                     & {\bf 83.0}                  & 57.0                       & {\bf 63.7} \\
		girl                   & 54.9                           & 87.7                    & {\bf 91.0}                  & 66.7                       & {\bf 76.6} \\
		hummingbird            & 52.0                           & 60.2                    & {\bf 75.0}                  & {\bf 65.2}                 & 64.5       \\
		monkey                 & 65.0                           & {\bf 79.0}              & 71.0                        & 80.5                       & {\bf 87.4} \\
		monkeydog              & 61.7                           & 39.6                    & {\bf 78.0}                  & 32.8                       & {\bf 51.4} \\
		parachute              & 76.3                           & {\bf 96.3}              & 94.0                        & 51.6                       & {\bf 88.4} \\
		penguin                & {\bf 18.3}                     & ~~9.3                   & -                           & {\bf 71.3}                 & 50.9       \\
		soldier                & 39.8                           & 66.6                    & {\bf 83.0}                  & 69.8                       & {\bf 83.2} \\
		worm                   & 72.8                           & {\bf 84.4}              & 81.0                        & {\bf 50.6}                 & 37.9       \\  \midrule
		Average                & 53.5                           & 57.3                    & {\bf 80}                    & 61.4                       & {\bf 64.3} \\ \bottomrule
		\end{tabular}
	   }
\end{table}

{\bf Performance on SegTrack-v2:} 
To demonstrate the performance of our method on the low-resolution dataset, we report the comparison results of our method with several available ones. 
Compared to the high-resolution DAVIS-2016 dataset, 
it is more difficult to predict accurate object regions with pretrained object proposals model on SegTrack-v2 dataset, as illustrated in \fig \ref{fig_fail-segtrack}.
NLC achieves the best performance on this dataset. However, it is an offline method based on non-local consensus voting of short-term and long-term motion saliency.
Compared to the online method, our approach achieves better performance in most videos, as shown in \tab \ref{tbl_video-segtrack}.

\begin{table}[!t]
	\centering
	\caption{Video object segmentation results on FBMS-59 test set with mIoU metric.}	
	\resizebox{1.0\linewidth}{!}{$
	\begin{tabular}{c|c|c|c|c|c}
	\toprule
	FST~\cite{ICCV2013_Papazoglou} & STP~\cite{CVPR2013_Zhang} & NLC~\cite{BMVC2014_Faktor} & ACO~\cite{CVPR2016_Jang} & ARP~\cite{CVPR2017_Koh} & UOVOS      \\ \midrule
	55.5                           & 47.3                      & 44.5                       & 54.2                     & 59.8                    & {\bf 63.9} \\ 
	\bottomrule
	\end{tabular}
	$}
	\label{tbl_fbms}
\end{table}

{\bf Performance on FBMS-59:} 
To further demonstrate the effectiveness of our method, we report the mIoU on the FBMS-59 test set. The results presented in \tab \ref{tbl_fbms} are obtained from ARP. Without parameters fine-tuning, UOVOS still reports the best performance among the compared algorithms. Moreover, our algorithm achieves an absolute gain of $+4.1$ when compared to the offline method ARP.

\begin{figure*}[!t]
	\centerline{\includegraphics[width=1.0\textwidth]{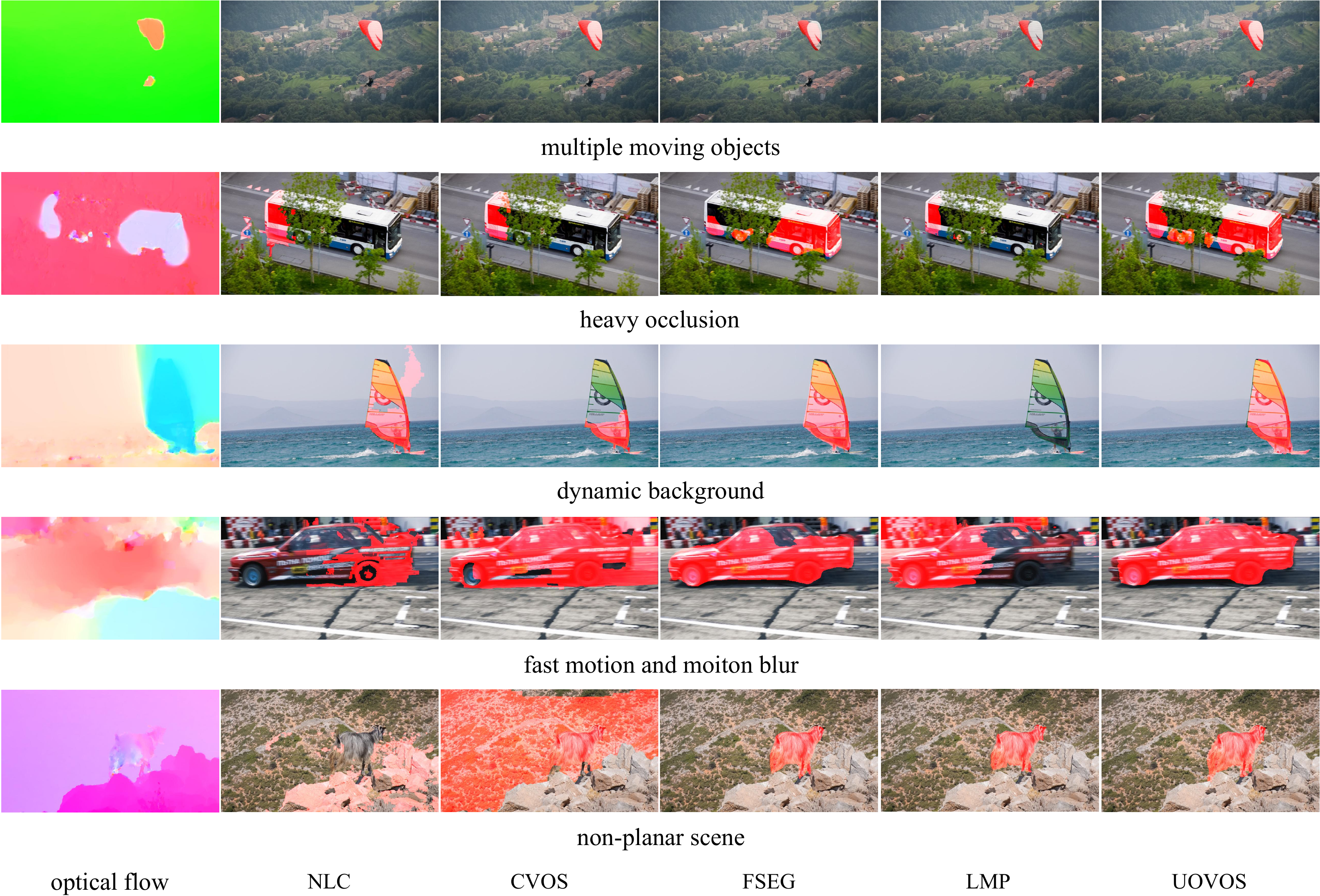}}
	\caption
		{
		Comparison on several challenging cases, which include multiple moving objects, heavy occlusion, dynamic background, fast motion and motion blur, and non-planar motion. 
		The compared algorithms are NLC~\cite{BMVC2014_Faktor} (offline), CVOS~\cite{CVPR2015_Taylor}, FSEG~\cite{CVPR2017_Jain} and LMP~\cite{CVPR2017_Tokmakov}.
		}
	\label{fig_comp}
\end{figure*}

\subsubsection{Qualitative Evaluation}
To qualitatively evaluate our method, 
we compare our method with several unsupervised offline and online methods on some challenging cases, 
including multiple moving objects, heavy occlusion, dynamic background, fast motion and motion blur, and non-planar scene.
For performance comparison, 
we compare our method with the offline method (\ie NLC \cite{BMVC2014_Faktor}), automatic initialization and tracking strategy based method CVOS \cite{CVPR2015_Taylor}, and two deep learning based methods (\ie FSEG \cite{CVPR2017_Jain} and LMP \cite{CVPR2017_Tokmakov}).
The segmentation results on the above scenarios are illustrated in \fig \ref{fig_comp}.
We analyze the results of each scenario as follows.

\textbf{Multiple moving objects:}
An unconstrained video often contains multiple moving objects and the proposed UOVOS is able to segment them automatically.
Similar to FSEG and LMP, 
for videos with multiple moving objects, we treat them as a single foreground.
As shown in the first row of \fig \ref{fig_comp},
our method is able to segment the two moving objects in this video.
For the offline method NLC,
the moving person is classified as background which may be due to the small region size of this person.
CVOS cannot automatically initialize the moving person, and thus failed to segment both of moving objects.
The appearance stream of FSEG is not reliable to extract the object regions in this frame and failed to segment the moving person.
Based on accurate motion segmentation and object proposals, 
LMP and UOVOS are able to successfully segment both objects.
More results on multiple moving objects segmentation are reported in \tab \ref{tbl_video-segtrack}, 
such as the \emph{bmx}, \emph{drift}, \emph{monkeydog} and \emph{penguin} videos.

\textbf{Heavy occlusion:}
Occlusion is a very challenging problem in VOS, which can cause disconnected for long-term trajectories generation and drift problem for tracking.
As shown in the second row of \fig \ref{fig_comp}, 
due to the disconnection trajectories caused by heavy occlusion,
some background regions are classified as foreground by NLC.
In addition, the segmentation is incomplete to cover the whole bus.
CVOS uses an automatic object initialization and tracking strategy,
and thus it suffers from the drift problem from tracking.
The segmentation result of CVOS is also incomplete.
LMP is learned on ground-truth of optical flow and motion segmentation of specific dataset and thus the performance of LMP is stable, such as the result shown in this frame.
FSEG can achieve better performance by fusing object proposals and motion segmentation in a unified framework and our method is slightly better than FSEG.

\textbf{Dynamic background:} 
Dynamic background regions are difficult to remove without prior knowledge about the object.
As shown in the third row of \fig \ref{fig_comp}, 
NLC and CVOS cannot get an accurate segmentation in this video.
LMP failed to segment the moving object in this video.
Because LMP adopts an end-to-end framework that learns the motion pattern from ground-truth optical flow and binary motion segmentation on the rigid scenes.
Thus, it is difficult to obtain accurate results when the motion is caused by non-rigid background (such as waving water).
Based on salient motion detection and robust object proposals, 
our approach achieves good segmentation results.

\textbf{Fast motion and motion blur:}
When a object moves fast, it leads to unreliable optical flow estimation and motion blur.
As shown in the fourth row of \fig \ref{fig_comp},
due to the fast car motion, 
the computed optical flow field is not accurate enough to indicate the moving car's region.
Therefore, the segmentation result of NLC is incomplete and CVOS contains too many background regions.
Similar to the dynamic background condition, LMP cannot obtain good segmentation when the computed optical flow field is not reliable.
Based on the proposed robust forward propagation refinement, our method achieves better performance than FSEG in this frame.

\textbf{Non-planar scene:}
Because of the nature of projecting a 3D world to a 2D plane (optical flow field),
it is difficult to distinguish the moving foreground from static background when the scene is non-planar. 
As shown in the last row of \fig \ref{fig_comp},
due to the lack of prior knowledge about the object,
the segmented foreground masks computed by NLC and CVOS are very different from each other,
and both methods fail to obtain reliable segmentation results.
With the help of robust object proposals, 
our method is able to obtain performance as good as that of FSEG and LMP.
\section{conclusion}
\label{sec:conclusion}

In this paper, we presented a new framework for the unsupervised online VOS problem.  Motivated by two key properties of moving objects, namely ``moving'' and ``generic'', we propose to apply salient motion detection and object proposals techniques for this challenging problem. Moreover, we designed a pixel-level fusion method and a forward propagation refinement strategy to improve the segmentation performance. Comprehensive experiments performed on three benchmark datasets demonstrates the effectiveness of our method. Without fine-tuning the pre-trained Mask R-CNN model, our method can outperform existing state-of-the-art methods by a large margin. Besides, we indetail analyzed the results and showed how the proposed method deals with some challenging scenarios. 

This work explores the potential of combining the salient motion detection and object proposal techniques for VOS. We hope that it can motivate more unsupervised online VOS studies on this new framework in the future.

\ifCLASSOPTIONcaptionsoff
  \newpage
\fi

\bibliographystyle{IEEEtran}
\bibliography{references}

\begin{IEEEbiography}[{\includegraphics[width=1in,height=1.25in,clip,keepaspectratio]{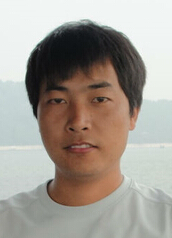}}]{Tao Zhuo}
	is currently a Research Fellow at the School of Computing, National University of Singapore. He received the M.E. and PhD degrees in Computer Science and Technology from Northwestern Polytechnical University, Xi’an, China, in 2012 and 2016, respectively. His research interests include image/video processing, computer vision and machine learning.
\end{IEEEbiography}

\begin{IEEEbiography}[{\includegraphics[width=1in,height=1.25in,clip,keepaspectratio]{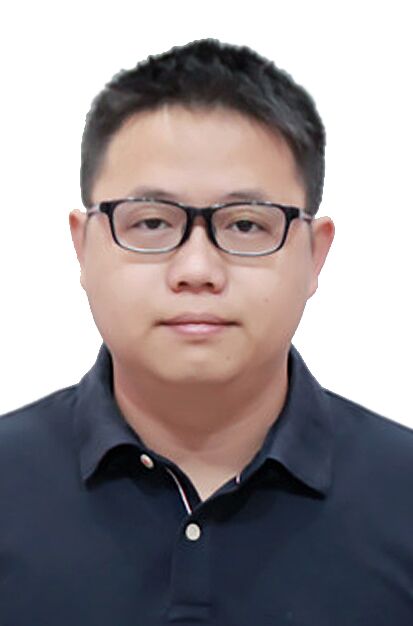}}]{Zhiyong Cheng}
	is currently a Professor with Shandong Artificial Intelligence Institute, Qilu University of Technology (Shandong Academy of Sciences). He received the Ph.D degree in computer science from Singapore Management University in 2016, and then worked as a Research Fellow in National University of Singapore. His research interests mainly focus on large-scale multimedia content analysis and retrieval. His work has been published in a set of top forums, including ACM SIGIR, MM, WWW, TOIS, IJCAI, TKDE, and TCYB. He has served as the PC member for several top conferences such as MM, MMM etc., and the regular reviewer for journals including TKDE, TIP, TMM etc.
\end{IEEEbiography}

\begin{IEEEbiography}[{\includegraphics[width=1in,height=1.25in,clip,keepaspectratio]{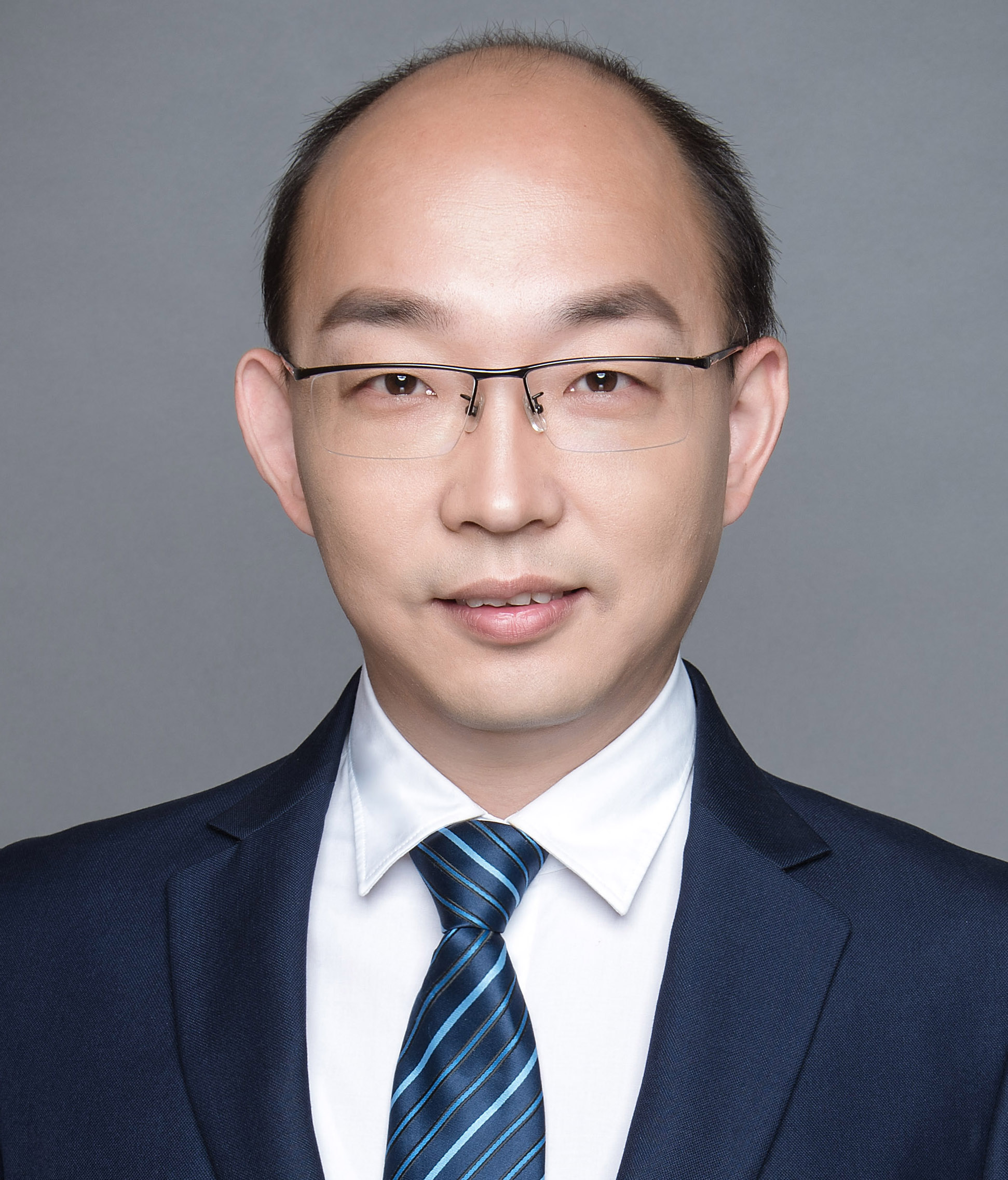}}]{Peng Zhang}
	is currently a full Professor in School of Computer Science, Northwestern Polytechnical University, China. He received the B.E. degree from the Xian Jiaotong University, China in 2001. He received his PhD from Nanyang Technological University, Singapore in 2011. His current research interests include object detection and tracking, computer vision and pattern recognition. He has published more than 80 high ranked international conference and journal papers and also has served as the technical committee in many international conferences and journals. He is a member of IEEE/ACM.
\end{IEEEbiography}

\begin{IEEEbiography}[{\includegraphics[width=1in,height=1.25in,clip,keepaspectratio]{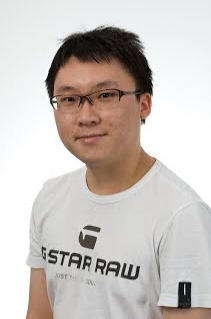}}]{Yongkang Wong}
	is a Senior Research Fellow at the School of Computing, National University of Singapore.
	He is also the Assistant Director of the NUS Centre for Research in Privacy Technologies (N-CRiPT).
	He obtained his BEng from the University of Adelaide and PhD from the University of Queensland.
	He has worked as a graduate researcher at NICTA's Queensland laboratory from 2008 to 2012.
	His current research interests are in the areas of Image/Video Processing, Machine Learning, and Social Scene Analysis.
\end{IEEEbiography}

\begin{IEEEbiography}[{\includegraphics[width=1in,height=1.25in,clip,keepaspectratio]{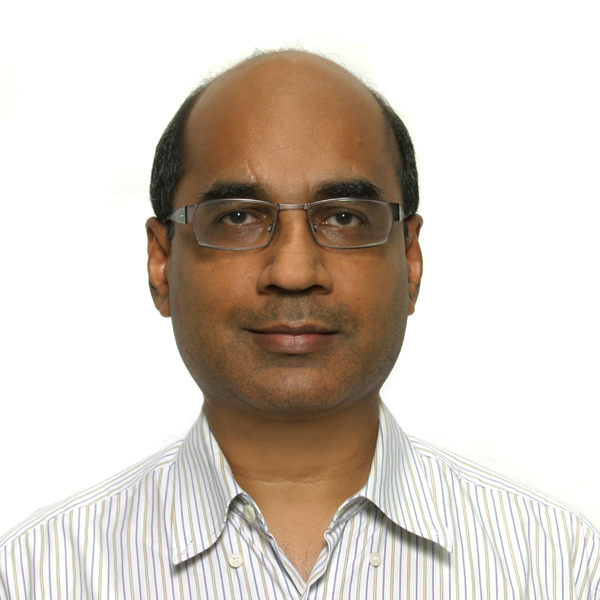}}]{Mohan~Kankanhalli}
	is the Provost's Chair Professor at the Department of Computer Science of the National University of Singapore.
	He is the Director of N-CRiPT and also the Dean, School of Computing at NUS.
	Mohan obtained his BTech from IIT Kharagpur and MS \& PhD from the Rensselaer Polytechnic Institute.
	His current research interests are in Multimedia Computing, Multimedia Security \& Privacy, Image/Video Processing and Social Media Analysis.
	He is active in the Multimedia Research Community and is on the editorial boards of several journals.
	Mohan is a Fellow of IEEE.
\end{IEEEbiography}

\end{document}